\def\arxivversion{1}
\documentclass{article}
\usepackage{iclr2027_conference,times}
\usepackage{microtype} 
\usepackage{graphicx}
\usepackage{float}
\usepackage[section]{placeins}   
\usepackage{amsmath,amssymb}
\usepackage{booktabs}
\usepackage{xcolor}
\usepackage{tikz}
\usepackage{colortbl}
\definecolor{rowmarkerfill}{HTML}{ECE9E5}
\definecolor{rowmarkeredge}{HTML}{B3A89A}
\definecolor{rowmarkerink}{HTML}{1A1A1A}
\DeclareRobustCommand{\rowmarker}[1]{\tikz[baseline=(n.base)]{\node[circle,
fill=rowmarkerfill,draw=rowmarkeredge,line width=0.65pt,inner sep=0pt,
minimum size=9pt,text=rowmarkerink,font=\fontsize{7}{7}\selectfont] (n) {#1};}}
\definecolor{muted}{HTML}{6F6F6F}
\definecolor{yes}{HTML}{5E8C57}      
\definecolor{part}{HTML}{B0904A}     
\definecolor{no}{HTML}{B0463A}       
\newcommand{\cy}[1]{\cellcolor{yes!22}#1}
\newcommand{\cp}[1]{\cellcolor{part!22}#1}
\newcommand{\cn}[1]{\cellcolor{no!18}#1}
\usepackage[colorlinks=true,allcolors=blue]{hyperref}

\graphicspath{{figures/}}   

\newcommand{\mrow}[2]{\shortstack[l]{\texttt{#1}\\[1pt]{\scriptsize\color{muted}#2}}}
\newcommand{\val}[2]{\shortstack{#1\\[1pt]{\scriptsize\color{muted}#2}}}

\newcommand{\app}[1]{App.~\ref{#1}}

\title{World Modeling in Transformers}
\ifdefined\arxivversion
\iclrfinalcopy
\newcommand{\projectglobe}{\tikz[baseline=-0.5ex,x=1em,y=1em,line width=0.45pt,line cap=round,line join=round]{%
  \draw (0,0) circle (0.44);
  \draw (0,0) ellipse [x radius=0.22,y radius=0.44];
  \draw (-0.381,0.22) -- (0.381,0.22);
  \draw (-0.44,0) -- (0.44,0);
  \draw (-0.381,-0.22) -- (0.381,-0.22);}}
\author{Pierre Beckmann\textsuperscript{1,2,3}\quad
Matthieu Queloz\textsuperscript{4}\quad
Andr\'{e} Freitas\textsuperscript{2,5}\\[5pt]
\normalfont\small\textsuperscript{1}EPFL, \textsuperscript{2}IDIAP Research Institute, \textsuperscript{3}MATS\\
\normalfont\small\textsuperscript{4}University of Bern, \textsuperscript{5}University of Manchester\\[6pt]
\normalfont\small{\hypersetup{urlcolor=black}\color{black}\projectglobe\hspace{0.4em}\textbf{Project website:} \url{https://bepierre.github.io/world-modeling/}}}
\let\submissionmaketitle\maketitle
\renewcommand{\maketitle}{\submissionmaketitle\lhead{Preprint}\renewcommand{\headrulewidth}{0.4pt}}
\hypersetup{pdftitle={World Modeling in Transformers},
  pdfauthor={Pierre Beckmann, Matthieu Queloz, Andr\'{e} Freitas}}

\else
\author{Anonymous}
\fi

\begin{document}
\maketitle

\begin{abstract}

Behavioral failures can make a transformer appear to lack a world model even when it has learned
faithful representations of its environment. We demonstrate this in TaxiGPT, a transformer trained on
random walks through Manhattan whose failures have been interpreted as evidence of an incoherent internal
map. Through mechanistic analysis and causal interventions, we show that the model represents
intersections and streets, tracks its position, and uses a goal compass to navigate. We trace its
failures to interference between superposed intersection features, which disrupts localization within the
internal map. Affordance packing, which groups representations of intersections with the same legal
moves, helps limit the consequences of these errors.  Finally, we propose mechanistic indicators that we
use to compare models and show that world-modeling capacities emerge at different stages of training.
Our findings motivate a shift from asking whether a model \emph{has a world model} to mechanistically
studying its \emph{world modeling}: the interacting capacities through which it represents its
environment and uses those representations to guide behavior.

\end{abstract}

\section{Introduction}

Among the more exciting promises of transformers is the prospect that entire ``world models'' might
emerge inside them just from training on sequences of data in a domain. Yet assessing whether a
transformer has recovered a faithful world model is more challenging than it first seems. A case in point
is what we dub ``TaxiGPT'': \citet{vafa2024worldmodel} trained GPT-2-style transformers from scratch on
taxi rides through Manhattan to see whether they would recover a faithful street map. The models were
trained to predict the taxi's next turn given a sequence of tokens encoding the taxi's origin,
destination, and preceding turns (``N NW NE E SW...''). The most accurate of these TaxiGPT models,
trained on random walks, was able to output legal turns 99\% of the time. But it operated in an
environment in which even a random guess had a non-zero chance of picking out a legal move. And when Vafa
et al. sought to reconstruct the maps implicit in the model's outputs, the map looked more like spaghetti
than like the street layout of Manhattan. The quality of outputs also deteriorated sharply when the
researchers intervened to force the taxi away from its destination three quarters of the time. Vafa et
al. concluded that their transformer was ``very far from recovering the true street map of New York
City'' \citep[p.~2]{vafa2024worldmodel}.

The difficulty, however, is that a model's behavior underdetermines the mechanisms that produced it.
Reconstructing the map implicit in its outputs cannot distinguish whether the model actually contains an
incoherent map, or whether it is the mechanisms that locate and guide the model within the map that
misfire.

Our mechanistic analysis of the TaxiGPT model trained on random walks reveals that the model in fact
harbors a \emph{highly faithful internal map} of Manhattan. The model accurately represents the
intersections and the streets connecting them. TaxiGPT keeps track of its location on the map using a
running state informed by a rolling window of previous positions. Building on causal investigations of
learned world representations \citep{li2023othello,spies2025}, we use targeted interventions to establish
that the model indeed exploits this map to navigate. Its representation of the current intersection
favors moves that are legal at that intersection.

To guide its selection between legal moves, it uses a \emph{goal compass} that encodes the destination's
direction and favors legal moves leading toward it. Ablating the compass preserves legal moves but
severely impairs the model's ability to reach distant destinations. This shows that simply identifying a
faithful internal map does not suffice to explain the model's navigational abilities. Mastering legal
turns within the street network and traveling toward a destination are distinct achievements calling
for different internal resources. There is more to world modeling than map recovery.

The question then becomes how a model using such a faithful map can nevertheless produce routes that
suggest an incoherent one. We find that TaxiGPT stores intersection representations in superposition.
Thousands of intersection features occupy a small subspace. Out of distribution, the signal specifying
the current position weakens and interference increases, allowing incorrect intersection features to
promote illegal moves. We trace the failures observed by \citet{vafa2024worldmodel} to this phenomenon.
Strengthening the correct position signal or suppressing interference substantially restores the model's
performance.

We further show that the model \emph{organizes this superposition by affordance}: intersections with the
same set of legal next moves tend to have nearby representations. This \emph{affordance packing} makes
many localization slips benign with respect to immediate move legality: a move that is legal at the
mistaken intersection is also legal at the true one. The rolling window of recent positions can then help
restore correct localization before the error propagates. \citet{vafa2025} also found evidence that
sequence models tend to group states with the same permissible next moves. \citet{prieto2026} showed that
overlapping representations of correlated features can helpfully reinforce one another. Our findings
illustrate a different benefit of affordance packing: it mitigates the damage caused by localization
slips.

These findings suggest that instead of looking for a single, self-contained ``world model'' at one layer,
we should look for multiple complementary \emph{world modeling} strategies: the mechanisms through which
a transformer represents environmental structure, tracks its situation within it, and uses that
information to guide prediction or action.

Finally, this decomposition yields \emph{mechanistic indicators of world-modeling capacities} that we
use to compare architectures, datasets, and training objectives, and to track the emergence of different
world-modeling capacities through training. This reveals that they do not emerge in lockstep: in the run
studied, legal-move and goal-direction representations mature before precise localization, while the
position code expands and becomes increasingly organized by affordance.

These results motivate a shift from identifying a \emph{world model} to explaining the processes of
\emph{world modeling}. A model is something a system \emph{has}; modeling is something it \emph{does},
often in several ways at once, and with uneven success. Learning the structure of an environment and
navigating it reliably are distinct achievements, which is why a faithful map can coexist with unreliable
navigation. Neither behavioral success nor representational fidelity on its own suffices to settle
whether and how a transformer models its world.

\section{World modeling in TaxiGPT}

\label{sec:map}

\subsection{A highly faithful internal map, stored in superposition}

We inspect GPT2-XL ($48$L $\times$ $1,600$ dims) trained on random walks over the map of Manhattan.

\textbf{Intersections are encoded.} We test two methods for extracting representations of the intersections,
or nodes, of the Manhattan map: linear probes and diff-means \citep{marks2023,li2023inference}.
Diff-means features score best. They decode most accurately at layer 18, where $99.6\%$ of intersections
are decoded with at least $90\%$ accuracy on held-out data (\app{app:intersection-layers}).
We next test whether these representations play a causal role in the model's predictions through a
minimal teleportation edit: we shift the taxi's encoded position to an intersection next to the goal
and check whether its next greedy prediction reaches the goal from that position. Using the layer 11
direction, this succeeds on at least one eligible scene for $99.3\%$ of testable nodes
(\app{app:teleport}). Finally, the intersection features encode Manhattan's spatial layout (\app{app:spatial}).

\begin{figure}[h]
\centering
\includegraphics{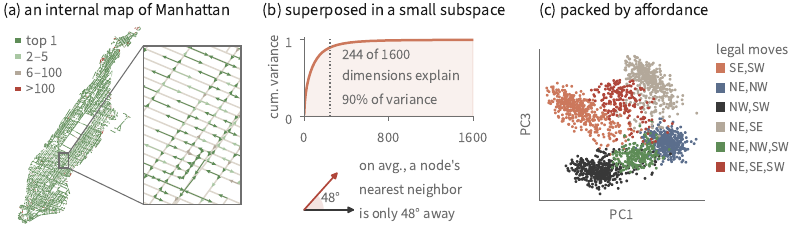}
\caption{\textbf{The map encoded inside TaxiGPT.} (a) The Manhattan map recovered by the street steering test:
we inject an intersection feature, feed a legal move, and measure which intersection feature grows most
afterward (color shows the correct next intersection's rank); (b) the intersection features are superposed in a small subspace of the residual stream; (c)
and are packed by affordance: intersections with the same legal next moves cluster together (main
affordance classes shown).}
\label{fig:hero-map}
\end{figure}

\textbf{Streets are encoded.} First, the intersection features themselves encode which moves are
legal from that point on the map. Applying the logit lens at layer~31 with a $1\%$
probability threshold recovers the exact legal-move set for $99.4\%$ of intersection features
(\app{app:reconstruction}). But this does not yet tell us whether the model encodes which intersections
are connected by each street. Two complementary tests provide evidence that the model encodes these too. The \emph{steering test} asks whether an intersection feature and a move activate the
feature of the correct next intersection. We inject an intersection feature, feed a move, and measure
which intersection feature grows most afterward. We average the other prompt states to avoid tying
the test to a particular route. The correct next intersection's feature grows most for $76.6\%$ of
tested streets and ranks among the top five for $93.3\%$ (Fig.~\ref{fig:hero-map}a). The \emph{probing
test} asks whether these transitions can also be recovered linearly from the model's residual-stream
state during rides. We train a separate linear probe for each move to predict the next intersection's
feature direction, then identify the intersection feature closest to the probe's output. This recovers
the correct successor for 89.2\% of legal state--move queries at held-out intersections
(\app{app:wiring}).

\textbf{The map is stored in superposition and packed by affordance.} The 4,516 intersection directions
of the position code occupy a small subspace of the 1,600-dimensional residual stream: 90\% of their
variance lies in 244 dimensions. The angle between an intersection direction and its nearest neighbor
averages 48.1$^\circ$ (Fig.~\ref{fig:hero-map}b). This superposition is
organized by affordance. We group intersections by their sets of legal next moves, yielding 104 groups,
and compute the mean feature vector for each group. For 91\% of intersections, the feature vector has
higher cosine similarity to its own group's mean than to any other group's mean. Thus, intersections
offering the same legal moves tend to have aligned features (Fig.~\ref{fig:hero-map}c).
Intersections that are close on the map also tend to have aligned features, though this
relationship is weaker (\app{app:superposition}).

\subsection{Localization and navigation}
\label{sec:localization-navigation}

Two world modeling capacities are linked to the internal map: the model \emph{localizes} itself on the
map (tracks where it is) and \emph{navigates} (works out which way to go from there).

\begin{figure}[h]
\centering
\includegraphics{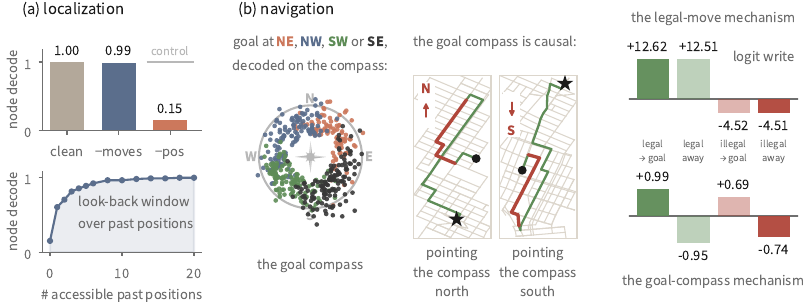}
\caption{\textbf{Using the map.} (a) \emph{Localization}: the model carries a running position, read from a
look-back window of past positions. Erasing past positions hurts current-intersection decode while
erasing the past moves does not (measured on rides of 27--34 moves). (b) \emph{Navigation}: a goal compass
encodes the bearing to the goal. It is also causal: clamping it north or south makes the taxi follow that
direction. Two mechanisms write into logits: a legal-move mechanism (activated by the intersection
feature) favors legal moves while the goal-compass mechanism favors goalward moves.}
\label{fig:hero-nav}
\end{figure}

\paragraph{Localization.}

TaxiGPT localizes by reading the active intersection features in the residual streams of several
past positions, which we call its \emph{look-back window}. It thus maintains a running position
estimate rather than recomputing its location from scratch using the sequence of moves
(Fig.~\ref{fig:hero-nav}a; \app{app:localize}).

\paragraph{Navigation.}

Two complementary mechanisms shape the model's move predictions (Fig.~\ref{fig:hero-nav}b). The active intersection feature
increases the logits of legal moves and decreases those of illegal ones. The \emph{goal compass}
increases the logits of moves toward the goal and decreases those of moves away from it. This compass
is a circular representation in the residual stream (similar to the circular features identified by
\citealp{engels2025,wurgaft2026}) that encodes the direction from the current intersection to the goal.
To identify it, we group rides into 16 bins by goal bearing, extract a difference-in-means direction
for each bin, and combine these directions to obtain the compass plane. Goal bearing is decoded best
at layer 16, with a median angular error of $18.1^\circ$ on held-out rides. Two interventions establish
the compass's causal role: steering it makes the model follow the selected direction over 12 moves,
with a median angular deviation of $18.1^\circ$, while ablating it preserves move legality but severely
impairs the model's ability to reach distant goals (\app{app:compass}).

\paragraph{Other mechanisms.}

The model uses an \emph{at-goal} feature to decide when to stop (\app{app:atgoal}). It also uses what we
call a \emph{commit-to-goal} feature. Because TaxiGPT is trained on random walks, it reproduces their
statistics: routes wander rather than head straight for the goal, and often overshoot it, looping back
before stopping. We find a single direction at layer 16 that controls this trade-off, dialing the model
between extreme random-walk behavior and shortest-path behavior (\app{app:nostop}).

These findings suggest that TaxiGPT satisfies two conditions for world representation discussed in the
philosophical literature: \emph{structural isomorphism}, supported by the encoding of intersections and
streets, and \emph{exploitation}, supported by the causal teleportation test \citep{shea2014,williams2026}.

\section{World modeling in superposition and resulting failure modes}

\label{sec:failures}

\begin{figure}[!ht]
\centering
\includegraphics{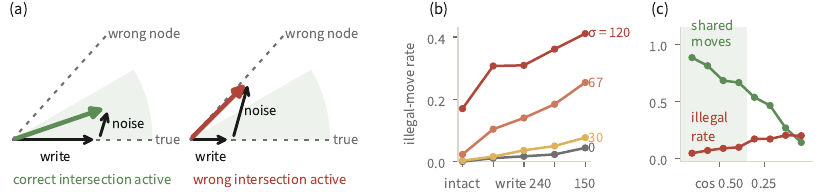}
\caption{\textbf{A weak write and noise in the position code are enough to cause off-graph moves.}
(a) When the true-position write is weak, noise can make a wrong intersection feature most active.
(b) We causally reproduce this phenomenon by weakening the write and adding noise to correctly localized states,
causing the model to emit illegal moves ($700$ stress rides with initially legal top moves,
depth $\geq 60$; edits at layer~18).
(c) The farther the wrong feature lies from the true direction, the fewer legal moves the two
intersections share and the more likely an illegal move becomes. The nearby region protected by affordance
packing is shaded green in (a) and (c).}
\label{fig:mechanism}
\end{figure}

We now explain why a model using such a faithful map can nevertheless produce the behavioral failures
reported by Vafa et al. We inspect their stress test, detour test, and compression metric. We show that
all three push the model into an out-of-distribution regime in which the current-position write grows
weaker and the noise in the position subspace higher. Because the map is stored in superposition, these
perturbations can activate a wrong node, making the next move illegal; though the affordance packing
limits how often this happens (Fig.~\ref{fig:mechanism}). We now look at each test in turn.

\subsection{The stress test}

\begin{figure}[!ht]
\centering
\includegraphics{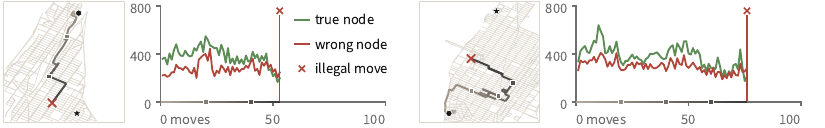}
\caption{\textbf{Two stress rides where a wrong node becomes most active, leading to an illegal move.} The shared fluctuations reflect how superposition works: when the true node is written more strongly, wrong nodes with superposed features also become more active.}
\label{fig:stress-examples}
\end{figure}

\begin{figure}[t]
\centering
\includegraphics{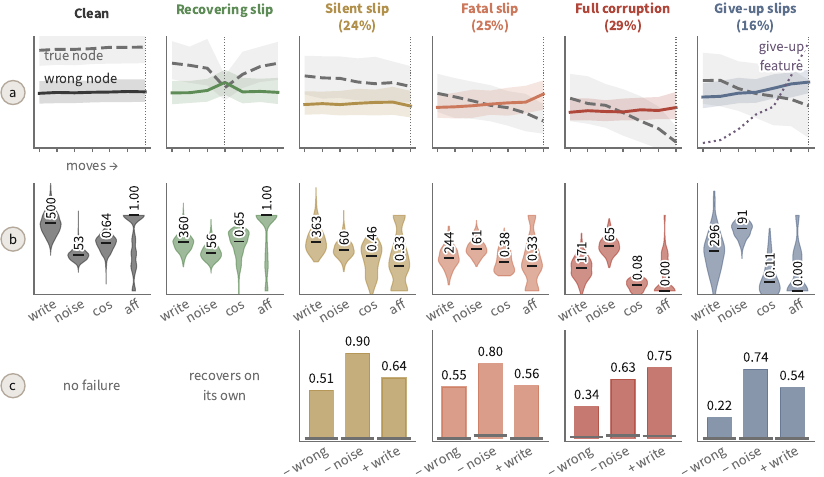}
\caption{\textbf{TaxiGPT's failure modes.}
\rowmarker{a} The first row shows how true-node (dashed) and wrong-node
(solid) activations change over successive moves, comparing clean rides, recovering slips and four failure modes.
\rowmarker{b} In these failure modes, weaker writes (write $\downarrow$)
and stronger noise (noise $\uparrow$) let the wrong node lie at a larger angle
(cos $\downarrow$) and share fewer legal moves (aff $\downarrow$), making illegal moves more likely.
\rowmarker{c} As a causal check, we remove the wrong-node activation
({\small$-$} wrong), remove position noise ({\small$-$} noise), or strengthen the true-position write
({\small$+$} write), and show the fraction of illegal-move probability removed
(gray marks: controls with edits of the same size along random directions).}
\label{fig:failures-hero}
\end{figure}

We use \emph{stress test} to refer to the evaluation underlying Vafa et al.'s reconstruction of the map
implicit in TaxiGPT's outputs. The model generates rides between sampled origin--destination pairs (with
temperature 1), and the resulting illegal moves are overlaid on Manhattan's true street map. These
sampled pairs place the model in a genuinely out-of-distribution regime: they are a median of $32$ moves
apart, whereas training rides start $9$ moves from their goal (\app{app:ood}). Because TaxiGPT replicates the meandering
of its training distribution, the goal is often unreachable within its budget of $99$ moves (after which it has no more trained position encodings). The model still generalizes quite well on
the stress test: it reaches the goal on $81\%$ of pairs and the taxi's current intersection can be
decoded from the model's internal activations with 99\% accuracy. But it does take an off-graph move on
$8.5\%$ of the rides. These failures stem from a weak write plus noise in position space, which can make a wrong node the most active and cause the model to emit an illegal move (Fig.~\ref{fig:stress-examples}). Given the amplitude of the perturbation, we can distinguish four categories of failure mode that lead to an illegal move (Fig.~\ref{fig:failures-hero}).

\paragraph{Recovering superposition slips.}

\emph{Superposition slips}, in which an incorrect intersection feature becomes most active, are mostly
benign. In benign cases, corruption of the position code is mild, combining a weak current-position write
(median 360; Fig.~\ref{fig:failures-hero}) with modest noise (median 56). The activated feature is close in direction to the correct
one (median cosine similarity 0.65), within the zone protected by affordance packing (median shared
affordance 1.0), so the next move remains legal. The look-back window over past positions then allows the
model to \emph{recover} from the slip (\app{app:collapse}).

\paragraph{Fatal slip ($25\%$).}

When the signal representing the true intersection weakens further (median strength 244), a less closely
aligned intersection representation can become active (0.38). The mistaken intersection shares fewer
legal moves with the true one (0.33), so a move that is legal there may be illegal at the taxi's actual
position. Removing the wrong node activation (the component orthogonal to the true node), removing the
position noise altogether (at L18), or strengthening the write all cause the illegal probability mass to
drop significantly ($0.55$ / $0.80$ / $0.56$).

\paragraph{Silent slip ($24\%$).}

Even when the correct node is the most active feature, wrong co-active nodes sometimes still leak illegal
moves into the logits through their affordances. They are almost entirely responsible: removing the
position noise drops the illegal probability by $0.90$.

\paragraph{Full corruption ($29\%$).}

When the write weakens further (median $171$), the corruption leaves the superposition regime: the
strongest competing node has little overlap with the true node (cosine $0.08$, shared affordance $0$).
Removing the top node therefore helps less ($0.34$); clearing all position noise ($0.63$) or rebuilding
the write ($0.75$) gives greater recovery.

\paragraph{Give-up slips ($16\%$).}

Deep in a ride, with the goal still far away, a give-up feature becomes active and promotes stopping (\app{app:giveup}). In
this regime, the residual is enlarged and noise in the position code can still produce slips. The write
is stronger than in full corruption (median $296$), but noise is also higher ($91$). Clearing this noise
reduces illegal probability mass by $0.74$.

These four categories account for $93.5\%$ of illegal moves. Of the rest, $6\%$ are low-mass unlucky draws:
the model places under $0.001$ total probability on off-graph moves (our marginal threshold), but
temperature-one sampling drew one anyway. The remaining $0.5\%$ fall outside these categories (\app{app:collapse}).

\begin{figure}[t]
\centering
\includegraphics{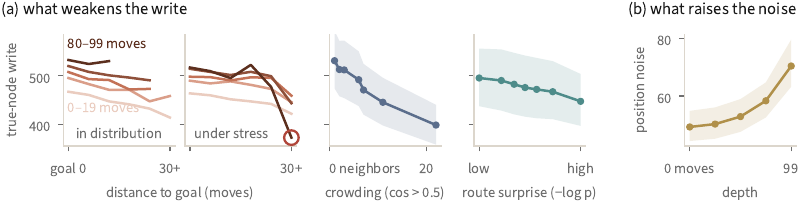}
\caption{\textbf{Factors affecting write strength and noise.} (a) In distribution, the write (measured at layer 18) falls
with distance to the goal but rises with depth. Under stress the pattern is broadly the same, except in
the out-of-distribution deep-and-far regime (red ring), where the write drops sharply. The write is also
weaker when the current intersection's representation overlaps with many others (crowding), and when the route is less predictable (route surprise). (b) Noise grows with one factor: depth. (Lines are
medians; bands are the interquartile spread.)}
\label{fig:write-drivers}
\end{figure}

\textbf{What shapes write strength and noise?} We observe that the write weakens with distance to the
goal, node superposition, and route surprise, and strengthens with depth
(Fig.~\ref{fig:write-drivers}). However, the stress test introduces a regime that training almost never
visits: \emph{deep-and-far}, where the model has taken at least $60$ moves and remains at least $20$ moves from
the goal ($0.1\%$ of in-distribution states against $5.4\%$ under stress). There, depth and distance
compound unexpectedly and the write becomes sharply weaker. Note that the distance-to-goal effect seems
to be intentional: swapping only the destination token in stress-ride states weakens the write in
$87\%$ of cases (median decrease: $59$) (\app{app:weakwrite}). The noise grows with depth.

The stress test thus reveals a \emph{misrecruitment} of the position code, driven largely by interference
between superposed intersection features. The failing world modeling capacity is primarily that of
\emph{localization}: the representations remain available, but the model struggles to recruit them
correctly under challenging conditions.

\subsection{The detour test and the compression metric}

\textbf{The detour test} (\app{app:detours}). At each move, the adversarial detour test overrides greedy
decoding with probability 0.75 to force the least-likely legal move. Once the remaining budget just
suffices to reach the goal, forcing stops and the model continues greedily. The model emits an illegal
move on $25.6\%$ of rides. We find that individual forced moves do not disrupt the internal
representations more than non-forced ones. However, because the least-likely legal move almost always
points away from the goal, repeated forcing pushes the ride into the deep-and-far regime and along routes
highly unlikely under the learned distribution. These conditions significantly weaken the
current-position write. The same four failure modes appear as in the stress test, with a shift toward
full corruption, and the same interventions improve move legality.

\textbf{The compression metric} (\app{app:compression}). Vafa's compression metric takes two same-length
routes (prefixes) that end at the same intersection with the same goal, samples 30 continuations
(suffixes) from one, and checks whether they remain likely under the other (probability $>\epsilon$). The
logic is that prefixes encoding the same state should support the same continuations. However, our
results point to the difficulty of the full ride formed by the prefix + suffix, rather than a localization mismatch
at the end of the prefixes. Indeed, both prefixes decode to the correct shared intersection in all 146
pairs we inspect, including the 94 whose continuations fail the test. What happens instead is that the
prefixes and suffixes form challenging rides that again enter out-of-distribution regimes, including
deep-and-far states, and exhibit the same four failure modes. To test whether compression tracks these
difficulties, we lengthen the suffixes by moving the goal farther away while keeping the prefix routes
fixed. Compression falls from 0.983 to 0.167 and illegal moves become more frequent, although both
prefixes still decode to the correct shared intersection in 99.8\% of conditions. Across the seven
distance bands, compression correlates strongly with illegal-move rate ($r=-0.957$), suggesting that
compression is sensitive to the same localization failures as the stress and detour tests.

Localization failures thus contribute to TaxiGPT's poor performance on the detour test and
compression metric. As a final confirmation, we continuously reinforce the correct position at L11 during
all tests (adding the correct intersection's diff-means vector after each move). This improves stress-test legality ($91.5\%\rightarrow 97.3\%$), detour success
($63.1\%\rightarrow 71.9\%$), and compression ($0.524\rightarrow 0.691$; \app{app:compression}).
This shows that the position code remains usable and effective; the difficulty lies in recruiting it
reliably.

\section{Mechanistic indicators of world modeling}

\label{sec:indicators}

\begin{figure}[!ht]
\centering
\includegraphics{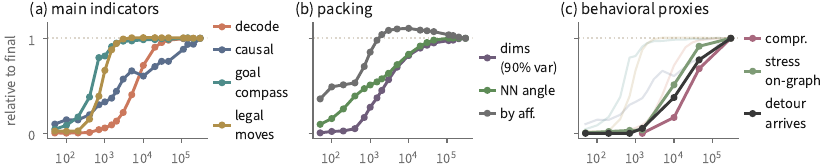}
\caption{\textbf{Emergence of world-modeling capacities through training} (RW\textperiodcentered  NTP, $384$d).
Navigation capacities (compass and legality) emerge first, localization (decode and causal) second (a).
The position code expands (nearest-neighbor angles grow); affordance packing rises early, then
relaxes as intersections differentiate (b). Mechanistic indicators can be compared with behavioral
proxies (c). Values are normalized to the final checkpoint.
(Protocol and controls in \app{app:growth}.)}
\label{fig:growth}
\end{figure}

We re-use the probing techniques developed for this case study as mechanistic indicators. This enables us
to compare world-modeling capacities across models, training data, and training regimes
(Table~\ref{tab:models}) and to track their emergence during training (Fig.~\ref{fig:growth}).

\begin{table}[h]
\caption{\textbf{Comparing world modeling capacities across architectures, data and training regimes.} For each
indicator, we select the layer where it scores best in its corresponding sweep. Rows are dataset \textperiodcentered{} objective: \texttt{SP} shortest paths, \texttt{NSP} noisy shortest paths,
\texttt{RW} random walks; \texttt{NTP} next-token prediction, \texttt{NextLat} next-latent prediction \citep{teoh2026nextlat}. (Protocol, controls, and training budgets in
\app{app:models}.)}
\label{tab:models}
\centering\footnotesize\setlength{\tabcolsep}{2pt}\renewcommand{\arraystretch}{1.1}
\resizebox{\textwidth}{!}{%
\begin{tabular}{@{}lccccccc|ccc@{}}
\toprule
& \multicolumn{5}{c}{\textbf{Map / localization}} & \multicolumn{2}{c}{\textbf{Navigation}}
& \multicolumn{3}{|c}{\textbf{Behavior}} \\
\cmidrule(lr){2-6}\cmidrule(lr){7-8}\cmidrule(lr){9-11}
& decode & causal & streets & super- & by & legal & goal & stress & detour & compr. \\
& & & probe & position & afford.? & moves & compass & test & test & \\
& {\scriptsize\color{muted}\% int $\ge$.9} & {\scriptsize\color{muted}telep $>$50\%}
& {\scriptsize\color{muted}held out}
& {\scriptsize\color{muted}angle / dims} & {\scriptsize\color{muted}own class}
& {\scriptsize\color{muted}\% int.} & {\scriptsize\color{muted}steer err}
& {\scriptsize\color{muted}on graph} & {\scriptsize\color{muted}success} & {\scriptsize\color{muted}score} \\
\midrule
\mrow{SP\textperiodcentered NTP}{12L$\times$768d$\times$12h}
 & \cn{\val{12.7\%}{L6}} & \cn{\val{14.9\%}{L10}} & \cn{\val{23.5\%}{L7}}
 & \cn{\val{35.9$^\circ$}{162d, L6}} & \cn{\val{41.0\%}{L6}} & \cn{\val{18.1\%}{L10}} & \cp{\val{25.8$^\circ$}{L12}}
 & \cp{71.7\%} & \cn{0.0\%} & \cn{.101} \\
\mrow{NSP\textperiodcentered NTP}{48L$\times$1600d$\times$25h}
 & \cn{\val{17.1\%}{L43}} & \cn{\val{16.9\%}{L46}} & \cn{\val{12.2\%}{L36}}
 & \cn{\val{38.0$^\circ$}{251d, L43}} & \cn{\val{41.1\%}{L43}} & \cn{\val{22.3\%}{L44}} & \cp{\val{23.2$^\circ$}{L47}}
 & \cp{74.7\%} & \cn{0.2\%} & \cn{.054} \\
\mrow{RW\textperiodcentered NTP}{48L$\times$1600d$\times$25h}
 & \cy{\val{99.6\%}{L18}} & \cy{\val{94.0\%}{L11}} & \cy{\val{89.2\%}{L15}}
 & \cp{\val{48.1$^\circ$}{244d, L18}} & \cy{\val{91.4\%}{L18}} & \cy{\val{99.4\%}{L35}} & \cy{\val{18.1$^\circ$}{L15}}
 & \cy{91.5\%} & \cp{63.1\%} & \cp{.524} \\
\midrule
\mrow{RW\textperiodcentered NTP}{48L$\times$384d$\times$8h}
 & \cy{\val{99.7\%}{L40}} & \cy{\val{88.5\%}{L31}} & \cy{\val{81.9\%}{L39}}
 & \cp{\val{47.1$^\circ$}{210d, L40}} & \cy{\val{88.2\%}{L40}} & \cy{\val{96.9\%}{L44}} & \cy{\val{17.3$^\circ$}{L43}}
 & \cy{96.7\%} & \cy{77.5\%} & \cp{.523} \\
\mrow{RW\textperiodcentered NextLat}{48L$\times$384d$\times$8h}
 & \cy{\val{99.9\%}{L36}} & \cy{\val{87.1\%}{L27}} & \cy{\val{94.8\%}{L44}}
 & \cp{\val{42.2$^\circ$}{183d, L36}} & \cy{\val{84.7\%}{L36}} & \cy{\val{99.0\%}{L44}} & \cy{\val{17.6$^\circ$}{L37}}
 & \cy{97.3\%} & \cy{78.5\%} & \cp{.556} \\
\bottomrule
\end{tabular}}
\end{table}

\textbf{Mechanistic indicators give a more fine-grained diagnostic than behavioral proxies.}
Behavioral scores alone cannot distinguish which world-modeling capacities are present in a model and which are not.
Mechanistic indicators can: Table~\ref{tab:models} shows, for example, that SP and NSP learn a causal
goal compass despite weak intersection decoding and street probing. We take the decode indicator to
be particularly useful because it tests both whether the model has representations that distinguish
intersections and whether it reliably recruits the correct representation during inference to localize. It reliably separates the
better- and worse-performing models (Table~\ref{tab:models}), follows a smooth sigmoid during training,
and closely tracks stress and detour performance (Fig.~\ref{fig:growth}). These results support its use as an
indicator in other world-modeling tasks, particularly when the decoded representations are also shown
to be causally used, as we establish here for the diff-means features.

\textbf{World-modeling capacities emerge at different stages of training.} In the run studied,
navigation capacities (legal moves and goal compass) mature before localization capacities (decode and
causal) (Fig.~\ref{fig:growth}). The diff-means intersection features encode legality before they
reliably distinguish intersections, suggesting that the model first groups intersections by legal-move
set, then differentiates them within each group. The goal compass provides a sense of direction before the model reliably distinguishes
the map's individual intersections and recruits their representations to guide its moves. These early spatial capacities may provide a foundation for learning
the map's precise relational structure. Stress and detour performance improves as this more precise structural
understanding develops. More generally, this suggests that we should expect world-modeling
capacities to sometimes develop unevenly, with some supporting the emergence of others.

\textbf{Better navigation need not imply a better internal map.} Smaller models achieve better
behavioral scores, yet the large RW model scores comparably on the mechanistic indicators and better on
the causal indicator. The behavioral gap appears to reflect differences in reliably using these
representations: the small RW\textperiodcentered  NTP model slips $3\times$ less and
RW\textperiodcentered  NextLat $6\times$ less (\app{app:models}). This also reinforces the distinction
between learning a faithful map and reliably locating oneself within it.

\textbf{Data, architecture, and training objective shape world modeling.} First, data appears to be the
most important lever: across all architectures, RW models have far better world-modeling representations
(Table~\ref{tab:models}). Given that the recorded failure modes emerge out of distribution, the data
could likely still be improved considerably (for instance by training on the deep-and-far regime).
Second, the architecture constrains world modeling: smaller RW models place their intersection
representations at later layers ($L27$--$L40$ versus $L11$--$L18$), suggesting that they compensate for a
narrower residual stream by spreading the necessary computation over more layers.
Third, the training objective can encourage the emergence of world-modeling capacities. NextLat, which
adds an objective of predicting the next latent state, achieves the best mechanistic scores and,
notably, encodes streets best. This fits the objective: streets determine
which intersection comes next after a move, so learning to predict the next latent state should
encourage the model to encode them.

\textbf{Affordance packing supports both early prediction and later robust world modeling.} During
training, the position code expands within the residual stream alongside improvements in world modeling
capacities (Fig.~\ref{fig:growth}). Better models likewise show less superposed features
(Table~\ref{tab:models}). The model learns to predict legal moves before it can reliably distinguish
individual intersections, and this early improvement accompanies increasing affordance packing.
Affordance packing thus first serves as an initial way to lower prediction loss. As training progresses,
individual intersections become distinguishable while the grouping largely persists. This is coherent
with our failure analysis: affordance packing helps protect against superposition interference; insofar
as this protection lowers training loss, it provides a further pressure to preserve (and potentially
reinforce) the packing. Our results thus show that an affordance bias (viewed critically by \citealt{vafa2025}) can coexist with real world modeling, and even help make it more robust.

\section{Related work}

\textbf{World representations and their use.} Behavioral evaluations reveal limitations in adaptive
planning \citep{momennejad2023}, state-consistent prediction \citep{vafa2024worldmodel}, and transfer
across tasks sharing the same underlying structure \citep{vafa2025}. Mechanistic interpretability studies have identified
world representations and state-tracking mechanisms in Othello \citep{li2023othello,nanda2023othello},
chess \citep{karvonen2024}, maze-solving transformers \citep{ivanitskiy2024,spies2025}, permutation
tasks (\citealp*{li2025statetracking}; \citealp{zhang2025finite}), and spatial language tasks
\citep{tehenan2025spatial,xia2026map}. Recent work formalizes the distinction between representing and
using world structure \citep*{li2025worldmodel}. \citet{lepori2026} find that models can learn
representations in context yet sometimes fail to use them when needed. Our analysis connects these approaches by showing how a causally used, faithful map can nevertheless
produce behavior that suggests an incoherent one.

\textbf{Packing in superposition.} Interference between superposed features can cause errors
\citep[e.g.,][]{stevinson2025}, and smart packing can limit the damage. Known strategies include antipodal packing
of features that do not co-occur \citep{elhage2022}, correlated packing of features that do
\citep{elhage2022}, which can be constructive \citep{prieto2026}, and hierarchical packing
\citep{park2025,bussmann2025}. We find affordance packing, where states that permit the same next actions
are placed close together, limiting the behavioral consequences of confusing their representations.

\section{Conclusion}

We investigate a model operating in a world with a finite number of states and deterministic transitions.
What additional capacities are needed for world modeling in more complex settings is a question for
future research. Nevertheless, TaxiGPT offers lessons about both the challenges of world modeling and
how we should assess it.

Superposition poses a challenge to reliable world modeling: interference between representations can
cause the model to apply its correct representations of the environment to the wrong situation. But
training may also find ways to limit this damage. Affordance packing helps preserve move
legality during localization slips: intersections with the same legal moves tend to have nearby
representations, so confusing them need not produce an illegal move. The look-back window of recent
positions can then support recovery.

Mechanistic interpretability helps distinguish explanations that behavior alone leaves unresolved. A navigation
error can stem from shortcomings in representing the environment, locating oneself within it,
determining which moves are legal, or selecting those legal moves that help one progress toward a
destination. By identifying how world-modeling capacities are implemented and interact, mechanistic analysis
allows us to trace where things go wrong beyond behavioral tests.

More broadly, our findings motivate a shift from asking whether a model has a world model to examining
its \emph{world modeling}. This means asking what environmental structures a model has learned, which
capacities recruit these structures to guide behavior, and under what conditions those capacities
work together reliably. Learning to accurately map out an environment is but a start; reliable world
modeling requires learning how to make good use of that map.

\subsection*{AI use statement}

We used LLMs throughout the research and writing process, including experiment design, coding,
figure preparation, and manuscript revision. In particular, rapid implementation of preliminary
experiments let us explore a wider range of hypotheses and identify promising signals for closer
investigation. We also continuously asked LLMs to identify errors in our claims and code. We take
responsibility for the final code, results, and manuscript.

\subsection*{Reproducibility statement}

Experimental details are provided in the appendix. Code and instructions for running
the experiments are available at
\ifdefined\arxivversion
\url{https://github.com/bepierre/world-modeling}.
Project website: \url{https://bepierre.github.io/world-modeling/}.
\else
\url{https://anonymous.4open.science/r/taxiGPT-ED9B/}.
\fi

\bibliographystyle{iclr2027_conference}
\bibliography{refs}

\begin{thebibliography}{26}
\providecommand{\natexlab}[1]{#1}
\providecommand{\url}[1]{\texttt{#1}}
\expandafter\ifx\csname urlstyle\endcsname\relax
  \providecommand{\doi}[1]{doi: #1}\else
  \providecommand{\doi}{doi: \begingroup \urlstyle{rm}\Url}\fi

\bibitem[Bussmann et~al.(2025)Bussmann, Nabeshima, Karvonen, and
  Nanda]{bussmann2025}
Bart Bussmann, Noa Nabeshima, Adam Karvonen, and Neel Nanda.
\newblock Learning multi-level features with matryoshka sparse autoencoders.
\newblock In \emph{International Conference on Machine Learning}, 2025.
\newblock arXiv:2503.17547.

\bibitem[Elhage et~al.(2022)Elhage, Hume, Olsson, et~al.]{elhage2022}
Nelson Elhage, Tristan Hume, Catherine Olsson, et~al.
\newblock Toy models of superposition.
\newblock \emph{Transformer Circuits Thread}, 2022.
\newblock arXiv:2209.10652.

\bibitem[Engels et~al.(2025)Engels, Michaud, Liao, Gurnee, and
  Tegmark]{engels2025}
Joshua Engels, Eric~J. Michaud, Isaac Liao, Wes Gurnee, and Max Tegmark.
\newblock Not all language model features are one-dimensionally linear.
\newblock In \emph{International Conference on Learning Representations}, 2025.
\newblock arXiv:2405.14860.

\bibitem[Ivanitskiy et~al.(2024)Ivanitskiy, Spies, R{\"a}uker,
  et~al.]{ivanitskiy2024}
Michael~I. Ivanitskiy, Alex~F. Spies, Tilman R{\"a}uker, et~al.
\newblock Structured world representations in maze-solving transformers.
\newblock \emph{arXiv preprint arXiv:2312.02566}, 2024.

\bibitem[Karvonen(2024)]{karvonen2024}
Adam Karvonen.
\newblock Emergent world models and latent variable estimation in chess-playing
  language models.
\newblock In \emph{Conference on Language Modeling}, 2024.
\newblock arXiv:2403.15498.

\bibitem[Lepori et~al.(2026)Lepori, Linzen, Yuan, and Filippova]{lepori2026}
Michael~A. Lepori, Tal Linzen, Ann Yuan, and Katja Filippova.
\newblock Language models struggle to use representations learned in-context.
\newblock In \emph{Proceedings of the 64th Annual Meeting of the Association
  for Computational Linguistics (Volume 1: Long Papers)}, pp.\  14841--14857,
  2026.
\newblock \doi{10.18653/v1/2026.acl-long.676}.
\newblock URL \url{https://aclanthology.org/2026.acl-long.676/}.

\bibitem[Li et~al.(2025{\natexlab{a}})Li, Guo, and
  Andreas]{li2025statetracking}
Belinda~Z. Li, Zifan~Carl Guo, and Jacob Andreas.
\newblock (how) do language models track state?
\newblock In \emph{Proceedings of the 42nd International Conference on Machine
  Learning}, volume 267 of \emph{Proceedings of Machine Learning Research},
  pp.\  34429--34452, 2025{\natexlab{a}}.
\newblock URL \url{https://proceedings.mlr.press/v267/li25r.html}.

\bibitem[Li et~al.(2023{\natexlab{a}})Li, Hopkins, Bau, Vi{\'e}gas, Pfister,
  and Wattenberg]{li2023othello}
Kenneth Li, Aspen~K. Hopkins, David Bau, Fernanda Vi{\'e}gas, Hanspeter
  Pfister, and Martin Wattenberg.
\newblock Emergent world representations: Exploring a sequence model trained on
  a synthetic task.
\newblock In \emph{International Conference on Learning Representations},
  2023{\natexlab{a}}.
\newblock arXiv:2210.13382.

\bibitem[Li et~al.(2023{\natexlab{b}})Li, Patel, Vi{\'e}gas, Pfister, and
  Wattenberg]{li2023inference}
Kenneth Li, Oam Patel, Fernanda Vi{\'e}gas, Hanspeter Pfister, and Martin
  Wattenberg.
\newblock Inference-time intervention: Eliciting truthful answers from a
  language model.
\newblock In \emph{Advances in Neural Information Processing Systems},
  2023{\natexlab{b}}.

\bibitem[Li et~al.(2025{\natexlab{b}})Li, Vi{\'e}gas, and
  Wattenberg]{li2025worldmodel}
Kenneth Li, Fernanda Vi{\'e}gas, and Martin Wattenberg.
\newblock What does it mean for a neural network to learn a ``world model''?
\newblock \emph{arXiv preprint arXiv:2507.21513}, 2025{\natexlab{b}}.

\bibitem[Marks \& Tegmark(2023)Marks and Tegmark]{marks2023}
Samuel Marks and Max Tegmark.
\newblock The geometry of truth: Emergent linear structure in large language
  model representations of true/false datasets.
\newblock \emph{arXiv preprint arXiv:2310.06824}, 2023.

\bibitem[Momennejad et~al.(2023)Momennejad, Hasanbeig, Vieira,
  et~al.]{momennejad2023}
Ida Momennejad, Hosein Hasanbeig, Felipe Vieira, et~al.
\newblock Evaluating cognitive maps and planning in large language models with
  {CogEval}.
\newblock In \emph{Advances in Neural Information Processing Systems}, 2023.

\bibitem[Nanda et~al.(2023)Nanda, Lee, and Wattenberg]{nanda2023othello}
Neel Nanda, Andrew Lee, and Martin Wattenberg.
\newblock Emergent linear representations in world models of self-supervised
  sequence models.
\newblock In \emph{BlackboxNLP}, 2023.
\newblock arXiv:2309.00941.

\bibitem[Park et~al.(2025)Park, Choe, Jiang, and Veitch]{park2025}
Kiho Park, Yo~Joong Choe, Yibo Jiang, and Victor Veitch.
\newblock The geometry of categorical and hierarchical concepts in large
  language models.
\newblock In \emph{International Conference on Learning Representations}, 2025.
\newblock arXiv:2406.01506.

\bibitem[Prieto et~al.(2026)Prieto, Stevinson, Barsbey, Birdal, and
  Mediano]{prieto2026}
Lucas Prieto, Melody Stevinson, Melih Barsbey, Tolga Birdal, and Pedro A.~M.
  Mediano.
\newblock From data statistics to feature geometry: How correlations shape
  superposition.
\newblock In \emph{International Conference on Learning Representations}, 2026.
\newblock arXiv:2603.09972.

\bibitem[Shea(2014)]{shea2014}
Nicholas Shea.
\newblock Exploitable isomorphism and structural representation.
\newblock \emph{Proceedings of the Aristotelian Society}, 114:\penalty0
  123--144, 2014.
\newblock \doi{10.1111/j.1467-9264.2014.00367.x}.

\bibitem[Spies et~al.(2025)Spies, Edwards, Ivanitskiy, et~al.]{spies2025}
Alex~F. Spies, William Edwards, Michael~I. Ivanitskiy, et~al.
\newblock Transformers use causal world models in maze-solving tasks.
\newblock \emph{arXiv preprint arXiv:2412.11867}, 2025.

\bibitem[Stevinson et~al.(2025)Stevinson, Prieto, Barsbey, and
  Birdal]{stevinson2025}
Melody Stevinson, Lucas Prieto, Melih Barsbey, and Tolga Birdal.
\newblock Adversarial attacks leverage interference between features in
  superposition.
\newblock \emph{arXiv preprint arXiv:2510.11709}, 2025.

\bibitem[Tehenan et~al.(2025)Tehenan, Moya, Long, and Lin]{tehenan2025spatial}
Matthieu Tehenan, Christian~Bolivar Moya, Tenghai Long, and Guang Lin.
\newblock Linear spatial world models emerge in large language models.
\newblock \emph{arXiv preprint arXiv:2506.02996}, 2025.
\newblock \doi{10.48550/arXiv.2506.02996}.
\newblock URL \url{https://arxiv.org/abs/2506.02996}.

\bibitem[Teoh et~al.(2026)Teoh, Tomar, Ahn, Hu, Pearce, Sharma, Krishnamurthy,
  Islam, Lamb, and Langford]{teoh2026nextlat}
Jayden Teoh, Manan Tomar, Kwangjun Ahn, Edward~S. Hu, Tim Pearce, Pratyusha
  Sharma, Akshay Krishnamurthy, Riashat Islam, Alex Lamb, and John Langford.
\newblock Next-latent prediction transformers learn compact world models, 2026.
\newblock URL \url{https://arxiv.org/abs/2511.05963}.

\bibitem[Vafa et~al.(2024)Vafa, Chen, Kleinberg, Mullainathan, and
  Rambachan]{vafa2024worldmodel}
Keyon Vafa, Justin~Y. Chen, Jon Kleinberg, Sendhil Mullainathan, and Ashesh
  Rambachan.
\newblock Evaluating the world model implicit in a generative model.
\newblock \emph{arXiv preprint arXiv:2406.03689}, 2024.

\bibitem[Vafa et~al.(2025)Vafa, Chang, Rambachan, and Mullainathan]{vafa2025}
Keyon Vafa, Peter~G. Chang, Ashesh Rambachan, and Sendhil Mullainathan.
\newblock What has a foundation model found? using inductive bias to probe for
  world models.
\newblock In \emph{International Conference on Machine Learning}, 2025.
\newblock arXiv:2507.06952.

\bibitem[Williams(2026)]{williams2026}
Iwan Williams.
\newblock Can structural correspondences ground real-world representational
  content in large language models?
\newblock \emph{Mind \& Language}, pp.\  1--19, 2026.
\newblock \doi{10.1111/mila.70018}.

\bibitem[Wurgaft et~al.(2026)Wurgaft, Rager, Kowal, Shyam, Feucht, Bhalla,
  Haklay, Bigelow, Sarfati, McGrath, Lewis, Merullo, Goodman, Fel, Geiger, and
  Lubana]{wurgaft2026}
Daniel Wurgaft, Can Rager, Matthew Kowal, Vasudev Shyam, Sheridan Feucht, Usha
  Bhalla, Tal Haklay, Eric Bigelow, Raphael Sarfati, Thomas McGrath, Owen
  Lewis, Jack Merullo, Noah Goodman, Thomas Fel, Atticus Geiger, and
  Ekdeep~Singh Lubana.
\newblock Manifold steering reveals the shared geometry of neural network
  representation and behavior.
\newblock \emph{arXiv preprint arXiv:2605.05115}, 2026.
\newblock \doi{10.48550/arXiv.2605.05115}.

\bibitem[Xia et~al.(2026)Xia, Chen, Wang, Zhu, Zhang, Chen, and
  Xiao]{xia2026map}
Sirui Xia, Aili Chen, Xintao Wang, Tinghui Zhu, Yikai Zhang, Jiangjie Chen, and
  Yanghua Xiao.
\newblock Can {LLMs} learn to map the world from local descriptions?
\newblock In \emph{Proceedings of the 64th Annual Meeting of the Association
  for Computational Linguistics (Volume 1: Long Papers)}, pp.\  2823--2845.
  Association for Computational Linguistics, 2026.
\newblock URL \url{https://aclanthology.org/2026.acl-long.128/}.

\bibitem[Zhang et~al.(2025)Zhang, Du, Jin, Fu, and Jin]{zhang2025finite}
Yifan Zhang, Wenyu Du, Dongming Jin, Jie Fu, and Zhi Jin.
\newblock Finite state automata inside transformers with chain-of-thought: A
  mechanistic study on state tracking.
\newblock In \emph{Proceedings of the 63rd Annual Meeting of the Association
  for Computational Linguistics (Volume 1: Long Papers)}, pp.\  13603--13621.
  Association for Computational Linguistics, 2025.
\newblock \doi{10.18653/v1/2025.acl-long.668}.
\newblock URL \url{https://aclanthology.org/2025.acl-long.668/}.

\end{thebibliography}

\appendix
\AddToHook{cmd/subsection/before}{\FloatBarrier}
\newcommand{\reviewnote}[1]{\textcolor{red}{[Review: #1]}}
\makeatletter
\setlength{\@fptop}{0pt}
\setlength{\@fpsep}{12pt}
\setlength{\@fpbot}{0pt plus 1fil}
\makeatother

\section{Intersection features}
\label{app:representations}

\subsection{Decodability}
\label{app:intersection-layers}
\begin{figure}[!htb]
\centering
\includegraphics{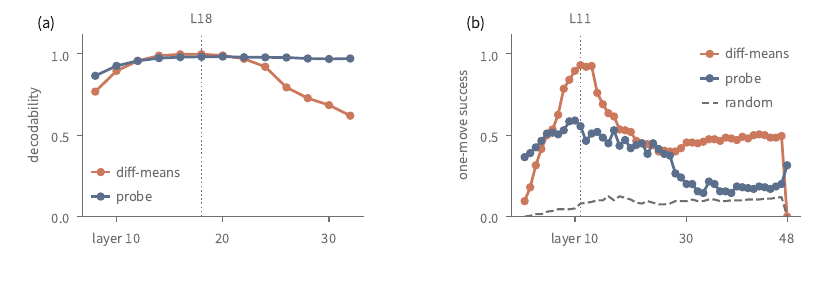}
\caption{\textbf{Reading and steering position across layers.} (a)~Held-out top-1 decode accuracy.
(b)~One-move success after minimal teleportation on $200$ scenes, with $\alpha=\beta=1.5$;
\emph{end} competes with the eight moves.
Dotted guides mark the selected reading layer (18) and steering layer (11).
The dashed curve is the random-node edit control for diff-means steering.}
\label{fig:readout-layers}
\end{figure}

\paragraph{Intersection representations identify the current position.}
We read the current intersection two ways: by its nearest centroid and with a linear
softmax probe. Both use the same training states per intersection and are tested
on held-out rides. An intersection's centroid $c_v$ is its mean residual; subtracting
the global mean gives its diff-means direction, $u_v=c_v-\bar c$.
Across $4{,}000$ rides ($201{,}779$ states), nearest-centroid accuracy peaks at $0.997$
at layer~18, versus $0.981$ for the probe. The probe remains accurate deeper in the
network while nearest-centroid decoding declines (Fig.~\ref{fig:readout-layers}a).
In a separate $6{,}000$-ride evaluation,
$99.6\%$ of the $4{,}497$ observed intersections decode at accuracy $\geq0.9$
(Table~\ref{tab:intersection-coverage}); the lowest accuracy is $0.778$.
We use layer~18 to read position.

\subsection{Causal interventions}
\label{app:teleport}
\begin{figure}[!htb]
\begin{minipage}[t]{0.48\linewidth}
\vspace{0pt}\centering
\includegraphics{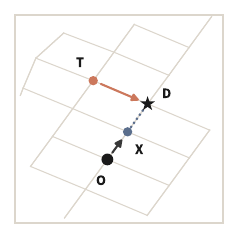}
\caption{\textbf{Minimal teleportation test.} After the taxi moves from $O$ to $X$, we edit its
position toward $T$. Success means its next greedy prediction is the orange move from
$T$ to the goal $D$, rather than the dotted blue move from $X$.}
\label{fig:teleport-scene}
\end{minipage}\hfill
\begin{minipage}[t]{0.49\linewidth}
\vspace{0pt}\centering
\makeatletter\def\@captype{table}\makeatother
\caption{\textbf{Teleportation success rate on $200$ scenes at layer~11.} Rows vary the
strength $\alpha$ of the added feature $u_T$; columns vary the strength $\beta$ of the
subtracted feature $u_X$. Bold marks the chosen pair.}
\label{tab:ab-sweep}
\small\setlength{\tabcolsep}{3pt}
\begin{tabular}{@{}lccccc@{}}
\toprule
$\alpha\backslash\beta$ & 1 & 1.5 & 2 & 2.5 & 3 \\
\midrule
1 & 0.765 & 0.825 & 0.690 & 0.595 & 0.490 \\
1.5 & 0.885 & \textbf{0.930} & 0.895 & 0.750 & 0.650 \\
2 & 0.885 & 0.910 & 0.900 & 0.860 & 0.755 \\
2.5 & 0.845 & 0.875 & 0.890 & 0.870 & 0.835 \\
3 & 0.775 & 0.850 & 0.870 & 0.865 & 0.830 \\
\bottomrule
\end{tabular}
\end{minipage}
\end{figure}

\paragraph{The minimal teleportation test.}
We test whether editing the model's position changes its next move as if it were at the
new intersection. We choose two intersections $X$ and $T$ one move from the same goal
$D$, each requiring a different move to reach it (Fig.~\ref{fig:teleport-scene}).
A prompt $[O,D,m_1]$ takes the taxi from origin $O$ to $X$. At the $m_1$ token,
we subtract the position feature of $X$ and add that of $T$:
$h\leftarrow h+\alpha u_T-\beta u_X$, once at layer~11, which gives the highest
teleportation success in the layer sweep (Fig.~\ref{fig:readout-layers}b).
We then take the highest-scoring
token among the eight moves and \emph{end}. The test succeeds if this move legally
reaches $D$ from $T$. We test both diff-means and probe directions.

\paragraph{Most intersection features causally guide the next move.}
For $99.3\%$ of the $4{,}202$ testable intersections, injecting its feature makes the
model choose the move from that intersection to the goal in at least one scene;
$94.0\%$ succeed in more than half their scenes (Table~\ref{tab:intersection-coverage}).
This evaluation covers all $24{,}877$ eligible scenes, using diff-means directions with
$\alpha=\beta=1.5$. The correct move's probability also increases in $97.9\%$ of scenes,
rising from $10.0\%$ to $56.9\%$ on average over the same eight moves plus \emph{end}.
On the $200$ scenes used to compare edit strengths, the model chooses the move from
$T$ to $D$ in $93.0\%$ of cases, against $0.0\%$ without an edit and $8.0\%$ with random
intersection features. Success remains high around strengths $1.5$--$2$
(Table~\ref{tab:ab-sweep}). Probe directions, scaled to the residual norm, reach at most
$59.0\%$ across their tested layers (Fig.~\ref{fig:readout-layers}b).

\begin{table}[!htb]
\centering\small
\caption{\textbf{Most intersection features are causal.} Teleportation succeeds for
$99.3\%$ of the $4{,}202$ testable intersections on at least one scene, and for $94.0\%$
on more than half. Decoding accuracy is reported below for the $4{,}497$ intersections
observed in held-out rides. Rows within each group overlap.}
\label{tab:intersection-coverage}
\label{tab:decode-bins}\label{tab:causal-bins}
\begin{tabular}{@{}lrr@{}}
\toprule
criterion & intersections & share \\
\midrule
teleport succeeds on any scene (L11) & 4,172 & 99.3\% \\
teleport succeeds on $>50\%$ of scenes & 3,951 & 94.0\% \\
teleport succeeds on every scene & 2,764 & 65.8\% \\
\midrule
decode accuracy $\geq0.9$ (L18) & 4,478 & 99.6\% \\
decode accuracy $=1$ & 4,026 & 89.5\% \\
\bottomrule
\end{tabular}
\end{table}

\FloatBarrier
\subsection{Spatial structure}
\label{app:spatial}
\begin{figure}[!htb]
\centering
\includegraphics{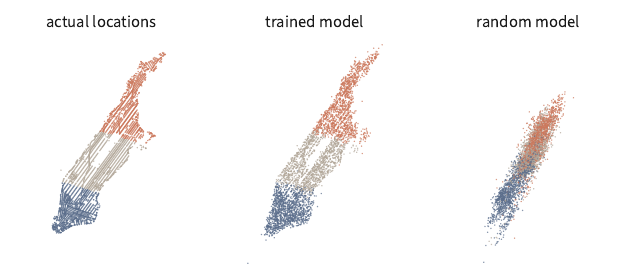}
\caption{\textbf{A linear readout of intersection features recovers Manhattan's spatial layout.} Actual locations
(left) and coordinates predicted from layer~6 features in the trained (middle) and
random (right) models, on the same scale. Colors mark three geographic bands along Manhattan's long axis,
kept fixed across panels. Held-out latitude/longitude $R^2$ is $0.988/0.988$ for the
trained model and $0.534/0.474$ for the random model. Maps include intersections used
to fit the readout.}
\label{fig:spatial}
\end{figure}

\paragraph{Intersection features encode geographic location.}
A linear readout predicts latitude and longitude for held-out intersections, reaching
$R^2=0.988$ for both coordinates at layer~6 (Table~\ref{tab:spatial}). We fit ridge
regression on $80\%$ of mapped intersections and test on the remaining $20\%$, using
mean-centered feature directions and standardized coordinates (penalty $100$, seed~0).
Fig.~\ref{fig:spatial} shows the predicted coordinates at layer~6. The same readout
is much less accurate on a randomly initialized model of the same architecture.
We do not show causal exploitation of this encoded spatial structure here;
evidence of causal use of spatial information comes from the goal compass
(Section~\ref{sec:localization-navigation}; \app{app:compass}).

\begin{table}[!htb]
\caption{\textbf{Coordinates are linearly readable from intersection features across layers.} Latitude/longitude
$R^2$ on the $20\%$ of intersections held out from fitting. The random-init control uses layer~6.}
\label{tab:spatial}
\centering\small\setlength{\tabcolsep}{4pt}
\begin{tabular}{@{}lccccccccc@{}}
\toprule
layer & 6 & 8 & 10 & 12 & 14 & 16 & 18 & 20 & random-init \\
\midrule
latitude & \textbf{.988} & .984 & .977 & .971 & .968 & .964 & .961 & .960 & .534 \\
longitude & \textbf{.988} & .985 & .979 & .975 & .972 & .969 & .966 & .965 & .474 \\
\bottomrule
\end{tabular}
\end{table}

\FloatBarrier
\subsection{Superposition and affordance packing}
\label{app:superposition}
\begin{figure}[!htb]
\begin{minipage}[t]{0.59\linewidth}
\vspace{0pt}
\makeatletter\def\@captype{table}\makeatother
\caption{\textbf{Intersection features overlap and group by legal moves.} Measurements
use layer~18 features. Affordance means the set of legal moves. Map-near and map-far pairs
are the closest and farthest fifths of the pooled sample of same-affordance pairs, ranked by geographic distance.}
\label{tab:superposition}
\centering
\small
\begin{tabular}{@{}p{0.73\linewidth}r@{}}
\toprule
quantity & value \\
\midrule
dimensions holding $90\%$ of the variance & 244 \\
participation ratio & 101 \\
mean pairwise $|\cos|$, features / random & $.073$ / $.020$ \\
mean nearest-neighbor cosine & $.668$ \\
mean within-affordance cosine & $.234$ \\
nearest affordance mean is the feature's own & $91.4\%$ \\
nearest feature shares affordance & $61.0\%$ \\
clustering AMI, affordance / space & $.574$ / $.192$ \\
same-affordance feature-neighbor is closer than a random one & $85.8\%$ \\
within-affordance cosine, map-near / map-far pairs & $.290$ / $.213$ \\
\bottomrule
\end{tabular}
\end{minipage}\hfill
\begin{minipage}[t]{0.38\linewidth}
\vspace{0pt}\centering
\includegraphics{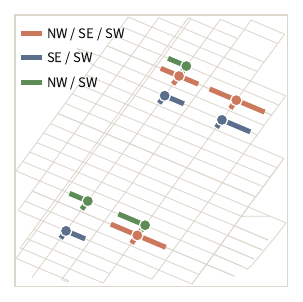}
\caption{\textbf{Superposed features at intersections with the same legal moves.}
Each color marks three such intersections. Within each group, two feature directions
have cosine similarity above $0.75$ to the third. Colored segments show their legal outgoing moves.}
\label{fig:affordance-map}
\end{minipage}
\end{figure}

\textbf{Intersection features occupy a small subspace and overlap.} At layer~18,
$244$ dimensions capture $90\%$ of the variance among $4{,}516$ intersection directions.
Their mean absolute cosine similarity is $.073$, compared with $.020$ for random
directions in the same $1{,}600$-dimensional space (Table~\ref{tab:superposition}).
We use unit-normalized diff-means directions and measure dimensionality from their
covariance eigenvalues $\lambda_i$; the participation ratio is
$(\sum_i\lambda_i)^2/\sum_i\lambda_i^2$.

\textbf{Intersection features are packed by affordance.} Feature clusters align
more closely with legal-move sets (the intersections' affordances) than with geographic
location. Fig.~\ref{fig:affordance-map} shows examples
of superposed features at different locations that share the same legal moves. We compare feature clusters with affordance
labels and geographic clusters using adjusted mutual information (AMI), with
$104$-cluster K-means for both features and coordinates. We also compare each feature
with normalized affordance-group means, which include the feature being scored.
For within-affordance cosine, we first average within each group of at least five
members, then average equally across groups.

\textbf{Superposition packing also favors neighboring intersections.} We sample pairs of intersections
with the same legal moves, using groups with at least ten mapped intersections. The geographically
closest fifth of these pairs has mean feature cosine similarity $.290$, compared with $.213$ for
the farthest fifth (Table~\ref{tab:superposition}).

\section{Street connectivity}
\label{app:connectivity}

\subsection{Legal moves}
\label{app:reconstruction}
\begin{figure}[!htb]
\centering
\includegraphics{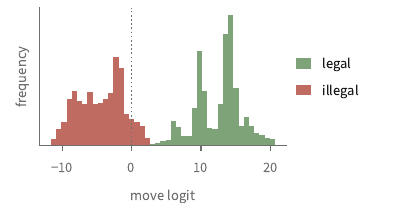}
\caption{\textbf{Move logits from centered intersection directions at layer~32, split by legality.}
We apply the logit lens to $c_n-\bar c$. Legal moves generally receive positive logits
and illegal moves negative logits.}
\label{fig:legality}
\end{figure}

\textbf{Intersection features encode which moves are legal.} Using the logit lens,
we classify moves with probability above $1\%$ as legal. This recovers the complete
legal-move set for $99.4\%$ of intersections at layer~31. We apply the model's final
layer normalization and output projection to each intersection centroid $c_n$, then
normalize over the eight moves. Fig.~\ref{fig:legality} illustrates the separation
between legal and illegal move logits.

\textbf{Legal moves are also readable from intersection features at earlier layers.}
A linear probe trained on $80\%$ of intersections achieves F1 $=0.989$ on the held-out
$20\%$ at layer~18 (Table~\ref{tab:reconstruct}). We use ridge regression with a fixed
split (seed~0).

\begin{table}[!htb]
\caption{\textbf{Legal-move F1 by layer.} The linear probe is fitted on training intersections and scored on
held-out intersections; logit-lens uses the model's output projection. Each entry uses
the threshold maximizing F1 on its reported scores. These scores pool intersection--move
pairs, rather than requiring the complete legal-move set to be correct.}
\label{tab:reconstruct}
\centering
\small
\begin{tabular}{lccccccc}
\toprule
layer & 18 & 20 & 24 & 28 & 32 & 36 & 44 \\
\midrule
linear probe & \textbf{.989} & .997 & .998 & .999 & .999 & .999 & .999 \\
logit-lens   & .937 & .962 & .993 & .999 & .999 & .999 & .999 \\
\bottomrule
\end{tabular}
\end{table}

\FloatBarrier
\subsection{Where the streets lead}
\label{app:wiring}
\begin{figure}[!htb]
\centering
\includegraphics{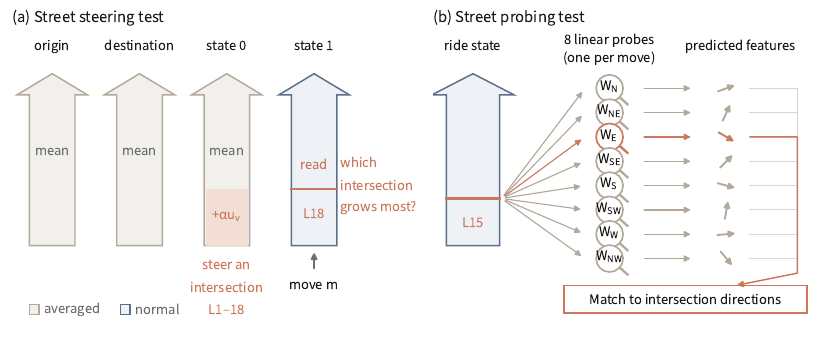}
\caption{\textbf{Two tests of street connectivity.} (a) The steering test asks whether
an intersection feature and a move activate the feature of the correct next intersection.
We inject an intersection feature into the averaged state~0 and feed move $m$, then measure which
intersection feature grows most at state~1. The other prompt states are averaged (beige)
to avoid tying the test to a particular ride; state~1 evolves normally (blue).
(b) The probing test asks whether street connections can be read from the model's state
during an ordinary ride. Eight linear probes, one per move, predict the feature of the
intersection that move would reach. We match each output to the closest intersection
feature and test on held-out source intersections.}
\label{fig:street-tests}
\end{figure}

\begin{figure}[!htb]
\centering
\includegraphics{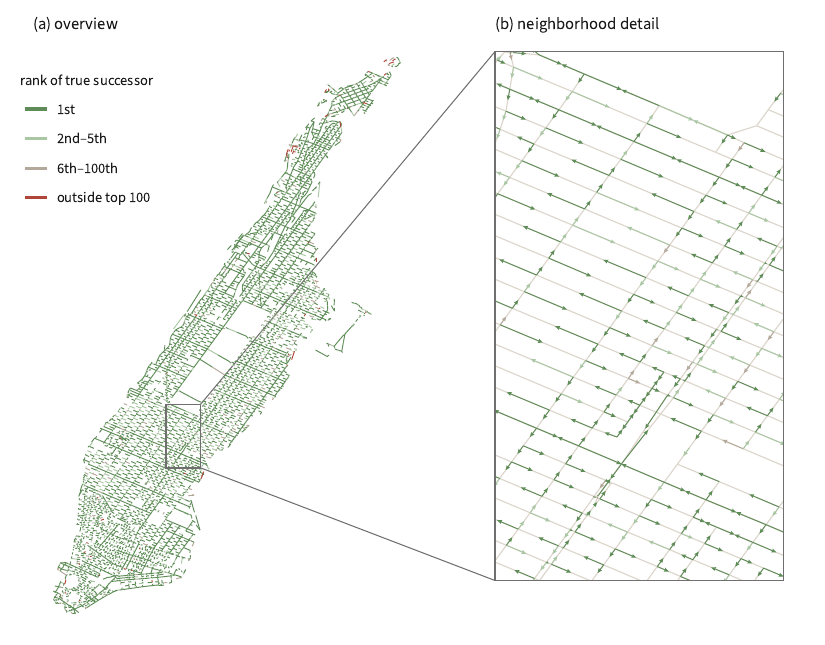}
\caption{\textbf{Street steering test.} Each tested move is drawn from its source intersection to
the street midpoint, so opposite directions can be colored separately. Color shows the correct next
intersection's rank among all candidates; the right panel enlarges the boxed area.}
\label{fig:internal-map}
\end{figure}

\textbf{Street steering test.} We inject an intersection feature,
feed a move, and measure which intersection feature grows most afterward
(Fig.~\ref{fig:street-tests}a). The other prompt states are averaged to avoid tying
the test to a particular ride. Across $8{,}934$ tested streets, the correct next
intersection ranks first in $76.6\%$ of cases and among the top five in $93.3\%$
(Fig.~\ref{fig:internal-map}). The four-token prompt contains an origin, destination,
position slot, and tested move. We replace the first three states with their layer-specific
means from held-out rides, then add the source intersection's normalized diff-means
direction at the position slot at every layer from 1 through 18. Its strength is $\alpha$ times the median
direction norm, with $\alpha\in\{0,1,2,4\}$. At the move token, we rank all $4{,}516$
normalized intersection directions by the increase in their layer-18 readout score
from $\alpha=0$ to $\alpha=4$. We also check whether the correct next intersection's
score increases monotonically across the four strengths.
The tested streets come from $3{,}820$ intersections with at least two tokenized exits
and available source and next-intersection features.

\textbf{Street probing test.}
We train eight linear probes, one per move, to read ordinary ride states and predict
the feature of the intersection that move would reach (Fig.~\ref{fig:street-tests}b). We match each prediction to the
closest of the $4{,}516$ intersection directions by cosine similarity. At layer~15,
accuracy is $89.2\%$ on queries from held-out source intersections, including moves not
taken during the ride. We hold out $25\%$ of source intersections, fit on $4{,}000$
training rides, and evaluate on $4{,}000$ held-out rides. The held-out sources' states
are excluded from fitting; their features remain among the candidate answers. Each probe
maps $h-\bar c$ to a next-intersection direction $c_v-\bar c$: we initialize it with
ridge regression, then fit with cross-entropy over cosine scores. We compute accuracy over
$113{,}998$ state--move queries covering $2{,}412$ distinct streets. The same
protocol predicting the current intersection reaches $99.3\%$; shuffled targets give $0.03\%$.

\FloatBarrier
\section{Localization and navigation}
\label{app:localization-navigation}

\subsection{The look-back window over past positions}
\label{app:localize}
\textbf{The model uses recent position features to locate itself.} Removing position
information from past states reduces current-position decoding accuracy from $99.8\%$
to $14.8\%$ on $27$--$34$-move rides. Preserving the most recent past position gives
$60.0\%$ accuracy, the last six give $91.7\%$, and the last twenty give $99.2\%$.
We replace each past move token's position-subspace component with that of its move
centroid to recompute its cached keys and values. The past forward pass stays fixed:
we do not propagate the edit through past tokens, and rerun the model only for the final token. The
subspace contains the top singular directions of the diff-means matrix accounting for
$90\%$ of its squared singular values. We preserve the orthogonal component, the origin
and destination states, and the final token, and decode at layer~18. The same edit in
a random subspace of matched dimension leaves accuracy at $99.8\%$; subtracting past
move diff-means directions leaves $99.0\%$. We leave the final move token unchanged, so the model can still combine that move
with a past position.

\begin{figure}[!htb]
\centering
\includegraphics{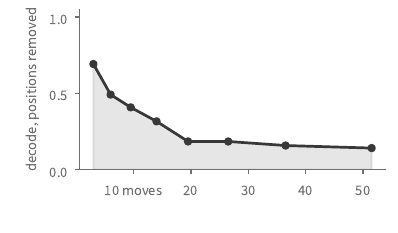}
\caption{\textbf{Without past positions, decoding worsens as the ride gets longer.}
We ablate past position features while preserving
the origin, destination and final token, then decode the current intersection at layer~18.}
\label{fig:localize-depth}
\end{figure}

\textbf{Past position features matter more on longer rides.} We repeat the ablation
at different ride lengths, removing all past position features. Current-position
decoding accuracy falls from $69.2\%$ at $2$--$4$ moves to $14.2\%$ at $43$--$60$ moves
(Fig.~\ref{fig:localize-depth}).

\FloatBarrier
\subsection{The goal compass}
\label{app:navigation}
\label{app:compass}
\begin{figure}[!htb]
\centering
\includegraphics{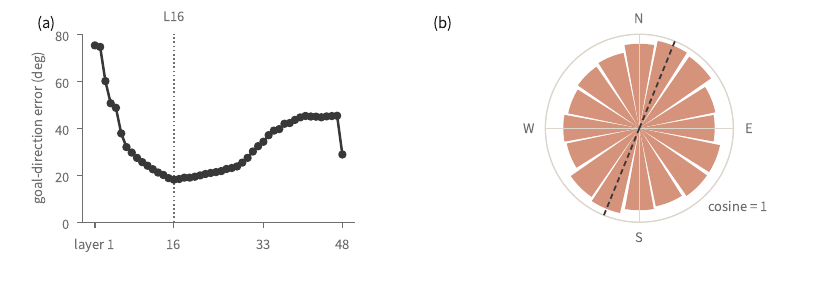}
\caption{\textbf{The goal compass encodes goal direction and guides travel.} (a) Error in
the goal bearing decoded from held-out ride states, by layer. (b) The model follows the
imposed direction across the $16$ tested bearings. Each wedge points in a direction
imposed on the compass. Its radius shows how closely the model travels in that
direction: mean cosine alignment, with $1$ for perfect alignment. Alignment is highest
near Manhattan's long axis
(dashed), estimated from the map coordinates.}
\label{fig:compass}
\end{figure}

\textbf{The goal compass encodes the direction to the goal.} We fit a two-dimensional
plane on training rides and use the angle within it to predict goal bearing on held-out
rides. The median decoding error is $18.1^\circ$ at layer~16
(Fig.~\ref{fig:compass}a). To fit the plane, we center training states within each
intersection to reduce the contribution of intersection identity, average them in
$16$ bearing bins, and combine the centered bin means $\mu_b$, with bin-center
bearings $\theta_b$, into two axes:
\[
v_{\cos}=\sum_{b=1}^{16}\cos(\theta_b)\mu_b,
\qquad
v_{\sin}=\sum_{b=1}^{16}\sin(\theta_b)\mu_b.
\]
We normalize each axis to unit length and decode bearing as
$\operatorname{atan2}(h^\top v_{\sin},h^\top v_{\cos})$, where $h$ is the residual
minus the fixed mean stored during fitting; evaluation does not use the current
intersection's identity. We fit on $6{,}000$ training
rides and evaluate on $3{,}000$ held-out rides, excluding states fewer than three moves
from the goal. Bearings use geographic coordinates with longitude corrected for latitude.
(Exploratory analysis with a sparse autoencoder revealed features sensitive to goal
bearing, motivating our extraction of the compass using differences of means.)

\textbf{The model follows the steered direction.} We set the compass to a target
bearing at each move, allow up to $12$ greedy legal moves, and measure the direction
from the ride's start to its endpoint. Across $16$ target bearings, the median error
is $18.1^\circ$ and the mean is $24.5^\circ$. Steering works across the tested bearings,
with the highest alignment near Manhattan's long axis (Fig.~\ref{fig:compass}b).
The edit removes the existing compass component with strength $\beta$ and adds the
target direction with strength $\alpha$, scaled by the remaining residual norm.
We orthonormalize the two axes before projecting out the compass component.
We use $\alpha=\beta=1$ at layer~15; Table~\ref{tab:compass-ab} compares strengths.
We retain $1{,}460$ rides with at least two moves and nonzero displacement, and compute
the mean and median of their individual angular errors. As a graph-only reference,
the best endpoint reachable in exactly $12$ legal moves has a median error of $1.0^\circ$.
The north/south examples in Fig.~\ref{fig:hero-nav}b use a stronger temporary edit
at layer~18 ($\alpha=16$, $\beta=1$): after $3$ initial moves, we steer for $12$ moves
and release the edit for up to $30$ further moves. Each phase starts a fresh prompt
at the current intersection with the same goal.

\begin{table}[!htb]
\caption{\textbf{Choosing the compass edit strength.} At layer~15,
$\beta$ scales removal of the existing compass component and $\alpha$ scales the added
target direction. Entries give mean $\cos(\text{heading}-\text{target})$: $1$ means aligned,
$0$ perpendicular, and $-1$ opposite. We use $60$ start--goal pairs and target bearings
$45^\circ$ and $225^\circ$; the $16$-bearing sweep uses $100$ pairs. Bold marks the chosen setting.}
\label{tab:compass-ab}
\centering
\small
\begin{tabular}{cccccc}
\toprule
 & $\beta{=}0$ & $\beta{=}0.5$ & $\beta{=}1$ & $\beta{=}1.5$ & $\beta{=}2$ \\
\midrule
$\alpha{=}0.25$ & .701 & .796 & .834 & .847 & .845 \\
$\alpha{=}0.5$  & .904 & .913 & .892 & .899 & .894 \\
$\alpha{=}1$    & .905 & .919 & \textbf{.920} & .922 & .917 \\
$\alpha{=}1.5$  & .843 & .846 & .857 & .857 & .826 \\
$\alpha{=}2$    & .505 & .513 & .512 & .484 & .434 \\
\bottomrule
\end{tabular}
\end{table}

\label{app:choose-move}
\textbf{Intersection features favor legal moves; the compass favors goalward moves.}
We read move preferences from intersection features using the logit lens, then compare
the effect of compass edits pointing toward versus away from the goal. Across
$2{,}817$ origin--goal prompts, centered intersection features at layer~32 give logits
near $+12$ for legal moves and $-4$ for illegal moves. For the compass, we set its
layer-16 component toward the goal, then away, and take the difference in output logits.
Among legal moves, the goalward-minus-awayward contrast is $0.11$ for the intersection
feature and $1.94$ for the compass intervention. The compass accounts for $95\%$ of
the sum of these two measured contrasts. We define goalward moves as those within
$60^\circ$ of the goal bearing and awayward moves as those beyond $120^\circ$, using
each move's average geographic direction and excluding the middle band. This also
assigns a direction to illegal moves. Compass edits use $\alpha=\beta=1$: the added
direction has the residual's norm after compass removal.

\begin{figure}[!htb]
\centering
\includegraphics{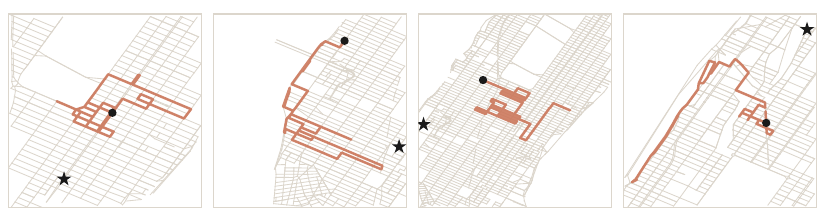}
\caption{\textbf{Without the compass, the model follows streets but misses the goal.}
Four selected rides after compass removal; circles mark origins and stars mark goals.
The corresponding baseline rides reach their goals.}
\label{fig:compass-wander}
\end{figure}

\textbf{Removing the compass impairs goal-reaching but preserves legal moves.}
We remove the compass plane at layer~18 and compare with removing a random plane of
the same dimension. For the immediate next move, compass removal changes the
highest-scoring legal move in $33.0\%$ of $2{,}500$ origin--goal prompts (random: $0.2\%$),
while the highest-scoring move remains legal in all three conditions. For complete
rides, goal-reaching falls from $94\%$ to $23\%$ in the farthest starting-distance bin
($18$--$32$ moves), while $99.5$--$99.7\%$ of moves remain legal across the distance
bins (Table~\ref{tab:compass-nav}). The random-plane control remains close to baseline.
We sample from the full vocabulary at temperature~1 for up to $128$ moves and count
a ride as successful if it visits the goal. Fig.~\ref{fig:compass-wander} shows four
examples that fail after compass removal.

\begin{table}[!htb]
\caption{\textbf{Compass removal reduces goal-reaching while moves remain legal.}
Starting distance is the minimum number of legal moves to the goal. The four distance
bins contain $31$, $50$, $115$, and $466$ tasks. Removing a random plane of the same
dimension serves as a control.}
\label{tab:compass-nav}
\centering
\small
\begin{tabular}{lcccc}
\toprule
start distance (moves) & 3--6 & 7--11 & 12--17 & 18--32 \\
\midrule
\emph{reaches goal:} & & & & \\
\quad with compass          & 1.00 & 1.00 & 0.99 & 0.94 \\
\quad compass ablated       & 0.68 & 0.54 & 0.38 & \textbf{0.23} \\
\quad random-plane ablated  & 1.00 & 1.00 & 0.98 & 0.94 \\
\midrule
\emph{moves that are legal:} & & & & \\
\quad compass ablated       & 0.995 & 0.997 & 0.996 & 0.995 \\
\bottomrule
\end{tabular}
\end{table}

\section{Other mechanisms}
\label{app:stopping}
\subsection{The at-goal feature}
\label{app:atgoal}
\begin{figure}[!htb]
\centering
\includegraphics{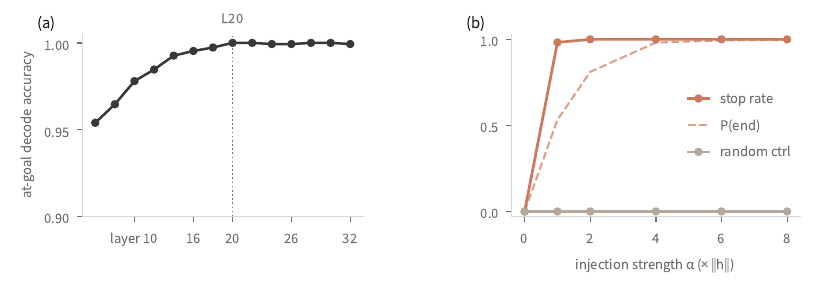}
\caption{\textbf{The at-goal feature detects arrival and can trigger stopping.} (a)~Held-out accuracy when the goal is set to the current intersection or a distant node.
(b)~Stop-token argmax rate and mean probability after injection at states away from the goal;
the random control shows its argmax rate.}
\label{fig:atgoal}
\end{figure}

\textbf{The at-goal feature fires when the model is at the destination.}
We compare the same ride state with two goals: its current intersection and a node at least
$15$ moves away. This holds the origin and move prefix fixed while changing whether the
ride is at its goal. We sample $1{,}500$ states from training rides.
We use half of these pairs to fit a difference-of-means direction and a midpoint
threshold, then test the readout on the other half. Accuracy reaches $100\%$ at layer~20, tied with several
later layers (Fig.~\ref{fig:atgoal}a). The model's mean $P(\text{end})$ is $0.55$ at the goal and below $0.001$ away.

\textbf{Adding the direction makes the model stop, even away from the destination.}
At $400$ evaluation states with a distant goal, we add the layer-20 direction at strength
$\alpha\lVert h\rVert$. The stop token becomes the highest-scoring token in $98.3\%$ of states
at $\alpha=1$ and $100\%$ at $\alpha=2$, versus $0\%$ without the edit. A random direction
with the same injection norm gives a $0\%$ stop rate across the tested strengths
(Fig.~\ref{fig:atgoal}b).

\textbf{The at-goal feature is distinct from a general stopping signal.}
We sample $3{,}000$ rides from origins $28$--$45$ moves from the goal and apply the fitted
layer-20 threshold at their final states. The feature fires at $93\%$ of successful stops,
but at $13\%$ of wrong-stop or generation-limit outcomes, despite high $P(\text{end})$ in
both groups (Table~\ref{tab:atgoal-dissoc}). The latter group has median length $99$ moves.
Illegal-move outcomes are excluded; rides without a stop are capped at $128$ moves.

\begin{table}[!htb]
\caption{\textbf{At-goal readout at final states of self-generated rides.} Generation-limit outcomes
reach the $128$-move cap without a stop token or illegal move.}
\label{tab:atgoal-dissoc}
\centering
\small
\begin{tabular}{lccc}
\toprule
outcome & feature fires & $P(\text{end})$ & $n$ \\
\midrule
stops at the goal & 0.93 & 0.80 & 2411 \\
wrong stop or generation limit & 0.13 & 0.89 & 383 \\
\bottomrule
\end{tabular}
\end{table}

\subsection{The commit-to-goal feature}
\label{app:nostop}
\begin{figure}[!htb]
\centering
\includegraphics{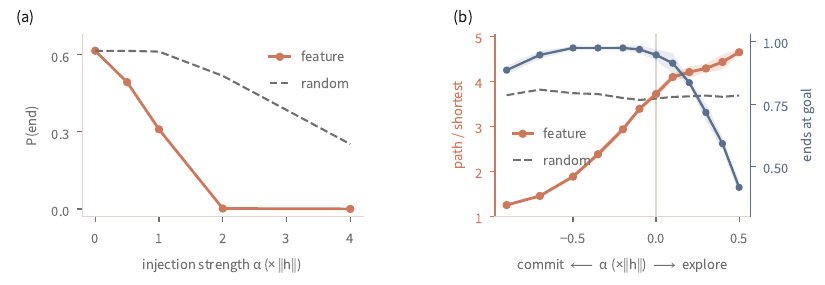}
\caption{\textbf{One direction controls stopping and how much the model meanders.} (a)~Stop probability after adding the continue direction
at terminal training-ride states. (b)~Path length relative to the shortest path (clay),
conditional on ending at the goal, and the fraction ending there (slate). Negative strengths
favor commitment; positive strengths favor exploration. Bands show one
standard deviation across three generation seeds; dashed lines show random-direction controls.}
\label{fig:nostop}
\end{figure}

\textbf{A single direction controls whether the model stops at the goal.}
From training rides, we collect states that end at the goal and states that pass through it
within the first $40\%$ of the ride, with at least five moves remaining. A difference-of-means
direction fitted on half of each class distinguishes the remaining states with $83.6\%$
accuracy at layer~16. We orient it toward continuing: adding it promotes exploration,
while subtracting it promotes commitment to the goal. Adding it at $300$ evaluation states where the training
ride ends reduces mean $P(\text{end})$ from $0.62$ to $0.002$ at $\alpha=2$, where the
injection norm is $\alpha\lVert h\rVert$. The random control gives $0.52$ at that strength
(Fig.~\ref{fig:nostop}a).

\begin{figure}[!htb]
\centering
\includegraphics{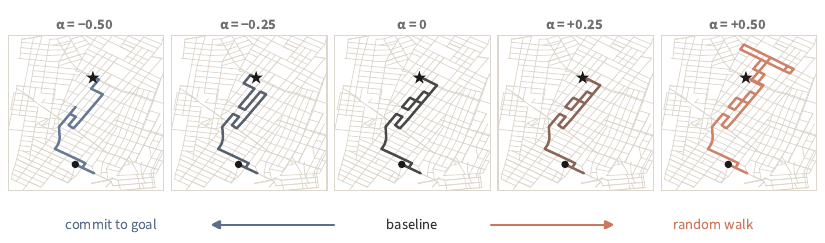}
\caption{\textbf{Steering changes the route on a fixed origin--goal pair.} One selected origin--goal pair at five steering strengths, with a $200$-move cap.
Circles mark the origin and stars the goal. Negative strength shortens this route;
positive strength adds loops.}
\label{fig:nostop_map}
\end{figure}

\textbf{The same direction also controls how much the model meanders.}
We test whether the same edit changes the route before arrival, adding it at each move
during full-vocabulary, temperature-1 generation on $120$ origin--goal pairs
$10$--$25$ moves apart, with a $128$-move cap. At $\alpha=-0.9$, the mean path length
among rides ending at the goal falls from $3.72$ to $1.26$ times the shortest path, while
the fraction ending there changes from $94.7\%$ to $88.6\%$. At $\alpha=0.5$, these values
are $4.64$ and $41.7\%$. We average over three generation seeds and compare with three
norm-matched random directions (Fig.~\ref{fig:nostop}b).
Fig.~\ref{fig:nostop_map} illustrates the route changes for one selected pair.

\begin{figure}[!htb]
\centering
\includegraphics{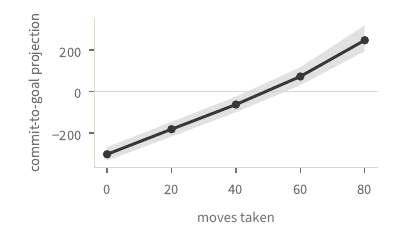}
\caption{\textbf{Commit-to-goal projection by depth on stress rides.} The line shows the median and
the band the interquartile range; higher values point toward committing to the goal.}
\label{fig:commit-budget}
\end{figure}

\textbf{Commitment to the goal increases on longer stress rides.}
For this readout, we fit the layer-16 direction on all collected stop and continue states,
then project states from $6{,}000$ self-generated stress rides onto it, with the sign reversed
so higher means more committed (Fig.~\ref{fig:commit-budget}). The projection increases
late in rides. This is consistent with the hypothesis that, as the model approaches
the roughly $100$-move training horizon, it commits more strongly to reaching the
goal within the remaining moves.

\section{The stress test}
\label{app:stress}

\subsection{The stress test is out of distribution}
\label{app:ood}

\begin{figure}[!htb]
\centering
\includegraphics{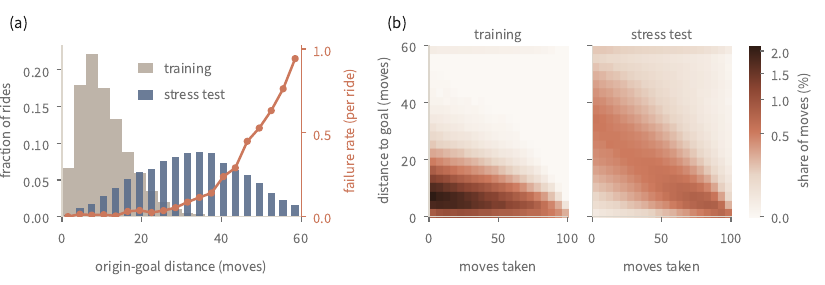}
\caption{\textbf{The stress test shifts both pair distances and visited states.}
(a)~Origin--goal distances in the released stress reference and held-out training-distribution
rides, with reference failure rate on the right axis. (b)~Visited-state occupancy for held-out
rides and our generated stress set. Heatmaps show depths up to $100$; their top distance bin
includes all distances of $57$ moves or more.}
\label{fig:ood}
\end{figure}

\textbf{Stress rides reach the deep-and-far regime.}
Among reachable pairs in the $6{,}400$ released rides, median origin--goal distance
is $32$ moves, versus $9$ in $50{,}000$ held-out training-distribution rides
(Fig.~\ref{fig:ood}a). The model therefore often remains far from the goal late
in a stress ride. The deep-and-far region, with at least $60$ moves taken and $20$
moves remaining, contains $5.4\%$ of stress states but only $0.1\%$ of held-out
states (Fig.~\ref{fig:ood}b). For this comparison, we use $8{,}000$ held-out rides
and $20{,}000$ generated stress rides, sampled from the full vocabulary at
temperature~1 with a $128$-move cap. We sample pairs from the released pool to
approximately match its distance histogram, treating reversed pairs as duplicates.

\textbf{Failures concentrate in the deep-and-far regime.}
In the released rides, failure increases with origin--goal distance
(Fig.~\ref{fig:ood}a). In our generated stress rides, the deep-and-far region
accounts for only $5.4\%$ of all visited states but $57.4\%$ of off-graph outcomes.
Failures thus occur disproportionately in a regime almost absent from training.

\subsection{Write strength and noise}
\label{app:weakwrite}

\textbf{Key information for Fig.~\ref{fig:write-drivers}.}
We measure $12{,}000$ stress rides at layer~18. After subtracting the global mean,
the write is the projection onto the unit true-intersection direction; noise is
the root-mean-square projection onto $96$ fixed sampled intersection directions.
Crowding counts other directions with cosine similarity above $0.5$ to the true
feature. Route surprise is the mean negative log-probability of the moves taken
so far, normalizing over the eight move tokens. For the goal-swap experiment,
we keep the origin and moves fixed in $800$ stress prefixes and replace only
the destination, choosing one $1$--$5$ or $30$--$69$ moves from the current
intersection. Moving the goal from near to far weakens the layer~18 write in
$86.5\%$ of cases, with a median paired decrease of $59.4$.
The substantial change from this single-token swap suggests a
learned adjustment of the position write, whose purpose remains unclear.
The swap also changes route probability, so it does not isolate distance from
other effects of goal conditioning.

\textbf{Why we interpret the remainder as noise.}
We test whether activity beyond the true-position write is concentrated on nearby
intersections or forms a stable part of the feature. After subtracting the write,
absolute projections toward intersections one to three moves away are only $25\%$
and $15\%$ larger than toward the rest of the map in $500$ clean and $500$ illegal
states. Comparisons across angles to the true feature, with random-vector controls,
also show broadly distributed activity with a modest preference for aligned directions.
For stability, we average the whole residual after subtracting the write across
visits to each intersection in $1{,}500$ held-out rides. Averaging $24$ visits reduces
its norm to $30\%$ of a single visit, versus $21\%$ for norm-matched isotropic noise.
The estimated fixed component accounts for $4.4\%$ of its energy (median over
intersections with at least $12$ visits, corrected for finite sampling), including
feature-estimation error. The remainder thus varies substantially across visits
and is not confined to nearby intersections, although it retains some structure.

\subsection{Illegal moves and recovery}
\label{app:collapse}

\textbf{Additional details for the main-text experiments.}
Table~\ref{tab:sorting} gives the ordered classification rule. Cases with total
illegal-move probability below $0.001$ are reported separately as low-mass unlucky
draws; all cases remain in the denominator.
All three rows of Fig.~\ref{fig:failures-hero} use the same failure cases:
rides with at least seven preceding moves, measured immediately before the sampled
illegal move, which need not be the highest-scoring move. We apply no further
sampling or illegal-probability cutoff for these rows. Clean and recovering
references are measured at their respective time-zero states. For the layer-18
repairs, we use full-vocabulary probabilities and report
$1-\sum P_{\rm illegal}^{\rm after}/\sum P_{\rm illegal}^{\rm before}$.
The wrong-node edit removes the strongest non-true direction after orthogonalizing
it against the true unit direction $u_T$. With $v=h-\bar\mu$ and $P$ projecting onto
the top $244$ singular directions of centered centroids, noise clearing subtracts
$Pv-\frac{(Pv)^\top u_T}{(Pu_T)^\top u_T}Pu_T$, preserving the true write and activity
outside the subspace. Write restoration raises the true-position write to $509$ if it is below that value. Controls use matched edit norms along random position-subspace
directions orthogonal to $u_T$. In Fig.~\ref{fig:mechanism}b, we cap the write at
$300$, $240$, $200$, or $150$ (plus an uncapped control), then add Gaussian noise
with coordinate standard deviation $0$, $30$, $67$, or $120$ in that subspace.
We add noise after capping the write, so the noise can also change the true-position
write. We record the highest-scoring move among the eight move tokens and the
intersection with the strongest activation.

\begin{figure}[!htb]
\centering
\includegraphics{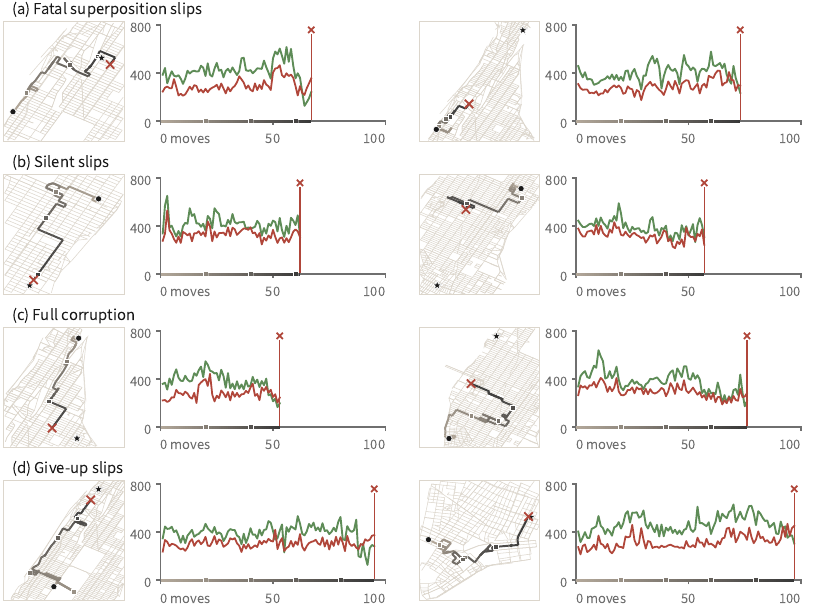}
\caption{\textbf{Stress trips illustrating four failure modes.}
Examples are selected near category medians at the final state. Maps mark the start
(black dot), goal (star), and illegal-move attempt (red cross). Matching colored squares mark every 20 moves on the route and time axis. Traces show the layer-18 true-position write (green) and
strongest wrong-node activation (red). True- and wrong-node activations often rise
and fall together. This is expected under superposition: intersection features have
overlapping directions, so strengthening the true-position feature also raises the
activation of wrong features that overlap with it.}
\label{fig:failure-examples-stress}
\end{figure}

\textbf{Individual stress rides illustrate the failure categories.}
Fig.~\ref{fig:failure-examples-stress} shows two trips per category. Both galleries
select distinct origin--goal pairs near the category medians of final-state write,
noise and wrong-node cosine, adding stopping activation for give-up slips. We
minimize the root-mean-square deviation scaled by each quantity's interquartile
range, subject to map readability.

\begin{table}[!htb]
\caption{\textbf{Failure categories, applied in the order shown.}
Here $v=h-\bar\mu$, $u_n=(\mu_n-\bar\mu)/\|\mu_n-\bar\mu\|$,
$T$ is the true intersection, and $L=\arg\max_n v\cdot u_n$.
The same direction scores define the strongest wrong node in the figures. A supplier is a non-true node among the $13$ highest direction scores
where the highest-logit illegal move is legal. Shares use all $1{,}681$ cases.}
\label{tab:sorting}
\centering\small
\begin{tabular}{@{}llr@{}}
\toprule
kind & condition & share \\
\midrule
low-mass unlucky draw & $P_{\rm illegal}<0.001$ & $6.0\%$ \\
give-up & else $v\cdot u_{\rm stop}\geq38.8$ & $16.2\%$ \\
silent slip & else $L=T$ and a supplier exists & $23.9\%$ \\
fatal slip & else $L\ne T$, $u_L\cdot u_T>0.2$ & $24.7\%$ \\
full corruption & else $L\ne T$, $u_L\cdot u_T\leq0.2$ & $28.7\%$ \\
unclassified & else $L=T$ without a supplier & $0.5\%$ \\
\bottomrule
\end{tabular}
\end{table}

\textbf{Steering toward a wrong intersection changes which move is selected.}
On $400$ clean deep states, we add a co-active wrong-node direction across
layers $10$, $12$, $14$, and $16$, scaled by each residual norm. The greedy move
becomes illegal at the true node in $18\%$, $42\%$, and $48\%$ of cases at total
strengths $0.5$, $1$, and $2$; in each case it is legal at the targeted node.
Steering toward the true node induces no illegal moves. The random-direction
control reaches $12\%$ at strength $2$.

\begin{figure}[!htb]
\centering
\begin{minipage}[t]{0.49\textwidth}\vspace{0pt}
\centering
\includegraphics{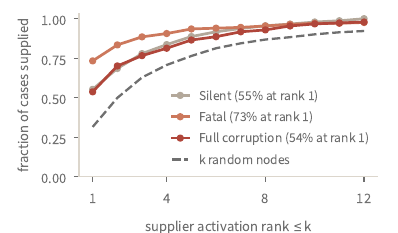}
\caption{\textbf{The co-active bag supports illegal moves more often than random nodes.}
For the bag of $k$ most active wrong-intersection features, curves show how often
the highest-scoring illegal move is legal at one of their intersections.
The dashed curve uses $k$ random nodes.}
\label{fig:bag-supplier}
\end{minipage}\hfill
\begin{minipage}[t]{0.49\textwidth}\vspace{0pt}
\centering
\includegraphics{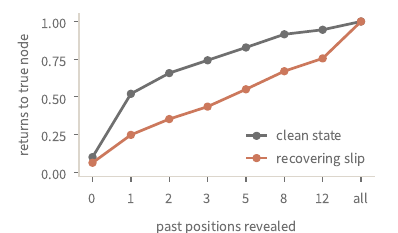}
\caption{\textbf{Recovering slips need more past positions than clean controls.}
Next-state accuracy after removing position information from older moves, on $400$
pairs matched exactly on depth and remaining goal distance. Both groups were
selected to decode correctly with the full history.}
\label{fig:recovery}
\end{minipage}
\end{figure}

\textbf{Illegal moves can be supported by a bag of co-active features.}
A corrupted position code can activate several wrong-intersection features at once.
We call the $k$ most active wrong features a \emph{bag of co-active features}, and
ask whether the highest-scoring illegal move is legal at any of their intersections.
These bags support the move more often than equally sized bags of random nodes,
especially for small $k$ (Fig.~\ref{fig:bag-supplier}). For silent slips, the most
active wrong feature alone supports the move in $55\%$ of cases, compared with
$32\%$ for a random node.

\textbf{Recovery after a slip relies on the look-back window over past positions.}
We select $400$ recovering superposition slip states where the decoded intersection is at least two moves from the true intersection and
returns to the true node on the next move, each paired with a clean control at exactly
the same depth and remaining goal distance. At that next move, we retain only the last $K$ past position codes and replace older
ones with their move-average components in the cached keys and values, without
propagating edits through earlier states. With $K=3$, accuracy is $43.5\%$ after
a slip and $74\%$ on clean controls (Fig.~\ref{fig:recovery}). Both reach
$100\%$ with the full history because we selected cases with correct decoding
at that next move. These percentages therefore do not measure how often slips
recover in general.

\subsection{Give-up slips}
\label{app:giveup}

\begin{figure}[!htb]
\centering
\includegraphics{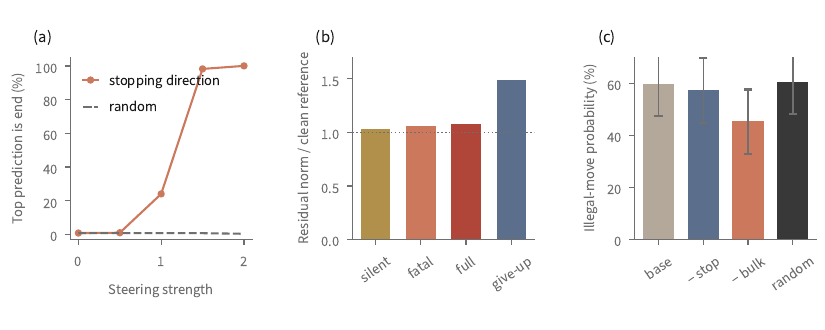}
\caption{\textbf{Give-up slips combine stopping activity with enlarged residuals.}
(a) Adding the give-up direction makes successful-ride states predict \emph{end}.
(b) Residual norms across failure categories, relative to clean states.
(c) On give-up prefixes, reducing excess residual activity lowers illegal-move probability
more than suppressing the stopping signal alone. Random edits match the reduction's
magnitude at each layer and position; error bars are $95\%$ bootstrap intervals over prefixes.}
\label{fig:giveup}
\end{figure}

\textbf{The give-up feature promotes stopping.}
We extract a direction by contrasting states where the model stops away from the goal
with late states from successful rides. We take the difference of their mean residuals
at layer~18, remove its position-subspace component, and normalize it. The give-up
direction is distinct from the at-goal direction (cosine $0.094$ at L18). To test whether
this direction promotes stopping, we add it to states from successful rides and measure
whether \emph{end} becomes the highest-scoring token. Adding the direction reliably
induces stopping, unlike a matched random direction (Fig.~\ref{fig:giveup}a).
We use $1{,}164$ states sampled every four moves from $60$ successful rides, adding
$\alpha/4$ times the residual norm at each of layers $10$, $12$, $14$, and $16$.

\textbf{Give-up slips have enlarged residuals and unusually high position noise.}
The stopping signal can be active while the model still produces an illegal move.
In these states, residual norms are larger than in the other failure categories
(Fig.~\ref{fig:giveup}b). The position code is also noisier, even though the true-position
write is stronger than in full corruption. Give-up slips thus combine the stopping signal
with a corrupted position code, which can still contain superposed wrong nodes.
For classification, we use a threshold of $5\%$ of the median projection onto
this direction in states where the model stops away from the goal.

\textbf{Reducing excess residual activity improves legality.}
On $32$ give-up prefixes, we compare removing the stopping signal with reducing excess
activity outside the position and stopping subspaces (Fig.~\ref{fig:giveup}c).
Removing only the stopping signal reduces stopping probability,
but barely changes illegal-move probability. For the residual reduction, we cap the
unprotected component at its median norm in healthy states, preserving the position and
stopping projections at each edit. This lowers illegal-move probability by $15.2$ percentage
points beyond matched random edits, while position noise remains high. Both interventions
act at every ride-state position across layers $14$--$47$. Local stopping directions are
fitted from $32$ wrong-stop and $24$ healthy states; layer~18 uses the classification direction.
We compare with three random directions, matching edit magnitudes at each layer and position.

\section{The detour test}
\label{app:detours}

\textbf{Test details.}
We generate detours using the reference rule: at each non-goal state,
with probability $0.75$, take the least-likely legal move among those that leave
the goal reachable within the remaining budget. Otherwise use the model's greedy
prediction. Success requires emitting \emph{end} at the goal. We use $10{,}000$
held-out origin--goal pairs and a $100$-move generation cap. The model stops at the goal on $63.6\%$ of rides, makes an illegal move on $25.6\%$, and
stops elsewhere on $10.7\%$. The main comparison uses the original benchmark
implementation; the analyses here use these $10{,}000$ generated rides.

\begin{figure}[!htb]
\centering
\includegraphics{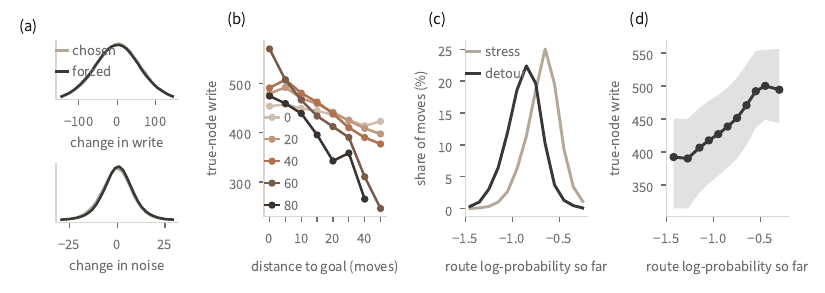}
\caption{\textbf{Individual forced moves do not immediately disrupt the position code, but repeated forcing creates deep-and-far, unlikely rides on which the position write weakens.}
(a) Move-to-move changes in write and noise after imposed versus self-chosen moves.
(b) Median write decreases with goal distance within depth bands.
(c) Detour routes are less likely than stress routes.
(d) Lower route probability accompanies weaker writes; shading shows the
interquartile range. Route log-probability is averaged over moves taken so far.
States above the give-up threshold are excluded throughout.}
\label{fig:detour-panels}
\end{figure}

\textbf{Individual forced moves do not cause a distinct drop in the position write.}
We compare move-to-move changes in write and noise after imposed and self-chosen
moves (Fig.~\ref{fig:detour-panels}a). The distributions are similar: forced
moves do not show a distinct immediate drop in write or rise in noise.
This comparison uses the moves as they occur, without matching the two groups.
All panels of Fig.~\ref{fig:detour-panels} exclude states above the
stress-fitted give-up threshold.

\begin{figure}[!htb]
\centering
\includegraphics{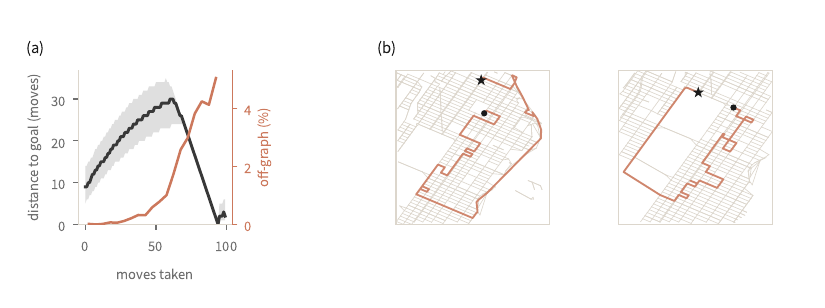}
\caption{\textbf{Detours first move away from the goal, then return.}
(a) Median goal distance over $10{,}000$ rides, with the interquartile range;
the clay line shows the percentage of ongoing rides going off-graph in each five-move bin (right axis).
(b) Two selected successful trajectories illustrating the reachability-constrained return;
dots mark origins and stars goals.}
\label{fig:detour-shape}
\end{figure}

\textbf{Repeated forcing creates deep-and-far, unlikely rides with weaker position writes.}
Goal distance rises to a median of about $30$ moves by move $60$, then falls as
the reachability constraint restricts the available detours
(Fig.~\ref{fig:detour-shape}). States with at least $40$ moves taken and at least
$20$ moves remaining account for $32.3\%$ of detour states, compared with $15.0\%$
of stress states and $0.4\%$ of held-out training-distribution states.
Within depth bands, the position write is weaker farther from the goal
(Fig.~\ref{fig:detour-panels}b). Detour routes also have lower mean move
log-probability than stress routes (panel c), and lower route probability
accompanies weaker writes (panel d). Repeated forcing thus brings rides into
both conditions associated with weak position writes.

\begin{figure}[!htb]
\centering
\includegraphics{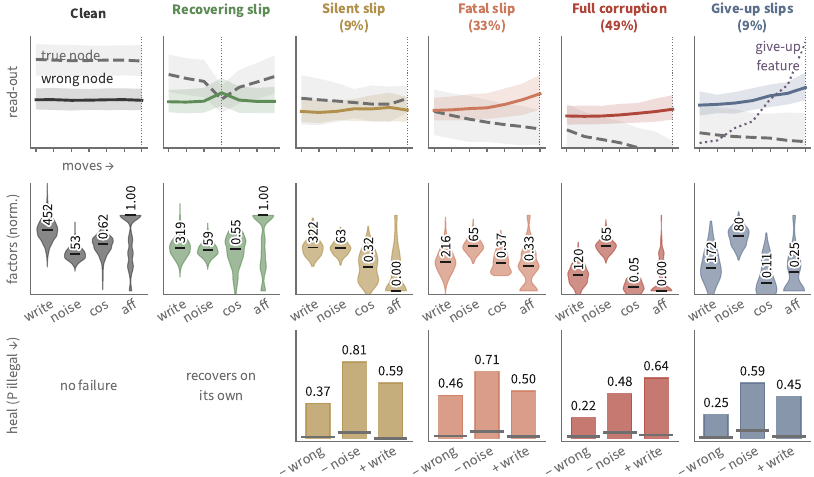}
\caption{\textbf{Detours exhibit the same failure modes: weak position writes and noise, with recovery under the same position edits.}
Columns compare failure categories and reference states. Rows show position readouts
around failure or recovery, scaled write/noise/cosine/affordance measurements, and the
fraction of illegal probability removed. Gray marks show matched random edits.}
\label{fig:hero-detour}
\end{figure}

\begin{figure}[!htb]
\centering
\includegraphics{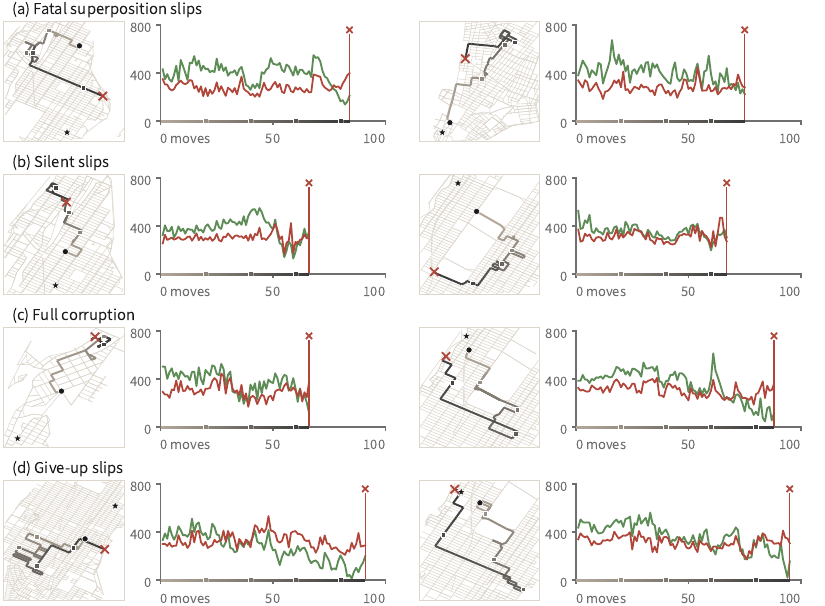}
\caption{\textbf{Detour trips illustrating four failure modes.}
Examples are near category medians at failure. Maps mark the start (dot), goal (star),
and illegal-move attempt (cross); squares mark every $20$ moves on the route and time
axis. Traces show layer-18 true-position write (green) and strongest wrong-node activation (red).}
\label{fig:failure-examples-detour}
\end{figure}

\FloatBarrier
\textbf{The same failure modes appear, with more full corruption.}
Applying the stress classifier gives $49.0\%$ full corruption, $32.7\%$ fatal
slips, $9.0\%$ silent slips, and $9.2\%$ give-up cases
(Fig.~\ref{fig:hero-detour}). Clearing position noise while preserving the
true-position write, or restoring the write, reduces illegal-move probability
in every category. Matched random edits remove at most $9\%$ of illegal
probability. Fig.~\ref{fig:failure-examples-detour} follows two individual
detour trips from each category.

\section{The compression metric}
\label{app:compression}

\begin{figure}[!htb]
\centering
\includegraphics{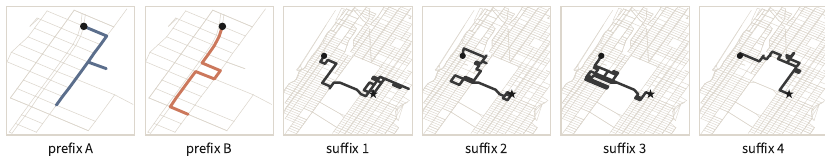}
\caption{\textbf{Two routes to one intersection, scored through their continuations.} Prefixes A and B have equal length and the same goal. Continuations sampled after A are scored after B; a trial passes only if every token in all $30$ continuations clears $\epsilon=0.01$. The drawing shows a selected subset.}
\label{fig:compression-schematic}
\end{figure}

\textbf{Metric details.}
Two equal-length prefixes end at the same intersection with the same goal.
We sample $30$ continuations after the first prefix and score them after the
second. A pair passes only if every scored token, including \emph{end}, has
probability above $\epsilon=0.01$ (Fig.~\ref{fig:compression-schematic}).
The compression score is the fraction of passing pairs. Prefix lengths vary;
continuations are sampled at temperature $1$ with an epsilon cutoff of $0.01$,
until stopping or a total sequence length of $128$ tokens.
The original-code score in the main comparison is $0.524$. The analyses below use supplementary runs.

\textbf{Both prefixes decode correctly even when compression fails.}
In the $146$ sampled-prefix trials inspected here, both prefixes decode to their
shared intersection in every pair, including all $94$ failing pairs. We read the
layer-18 mean-centered residual against normalized intersection directions at the
prefix endpoints, before generating the suffix. Table~\ref{tab:compression} orders
the two prefixes by write strength. Failing pairs tend to have more distant goals;
correct initial localization does not ensure that the model stays correctly localized
throughout the continuation.

\begin{table}[!htb]
\caption{\textbf{Prefix states in passing and failing compression pairs.} The $146$ sampled-prefix trials contain $52$ passing and $94$ failing pairs. Each pair is ordered by write strength. Entries are medians; parentheses give the first and third quartiles for goal distance and prefix length.}
\label{tab:compression}
\centering
\small
\begin{tabular}{lrrrr}
\toprule
at the end of a route & \multicolumn{2}{c}{pair passes} & \multicolumn{2}{c}{pair fails} \\
\cmidrule(lr){2-3}\cmidrule(lr){4-5}
 & weaker & stronger & weaker & stronger \\
\midrule
distance to the goal (moves) & \multicolumn{2}{c}{$27.5$ \,\small(21--40)} & \multicolumn{2}{c}{$54$ \,\small(41--67)} \\
length of the route (moves) & \multicolumn{2}{c}{$20.5$ \,\small(10--34)} & \multicolumn{2}{c}{$18.5$ \,\small(9--33)} \\
\midrule
write on the true intersection & $426$ & $457$ & $405$ & $438$ \\
noise in the position code & $51.3$ & $51.9$ & $51.0$ & $50.9$ \\
log-probability of the route & $-0.870$ & $-0.832$ & $-0.833$ & $-0.853$ \\
a wrong intersection is most active & $0\%$ & $0\%$ & $0\%$ & $0\%$ \\
\bottomrule

\end{tabular}
\end{table}

\begin{figure}[!htb]
\centering
\includegraphics{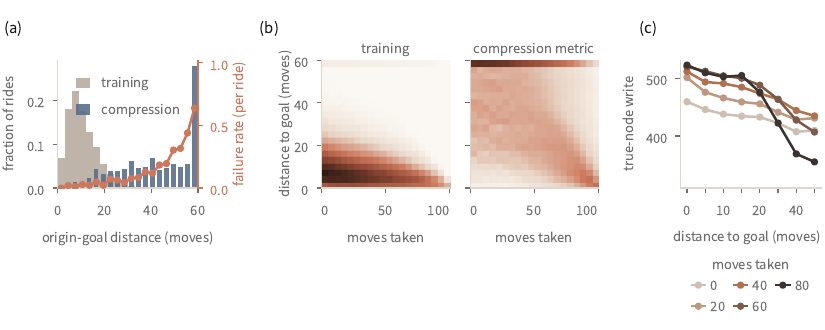}
\caption{\textbf{Compression generates rides far from the training distribution, where position writes weaken.}
(a) Origin--goal distances and off-graph ride rates. (b) Visited-state occupancy by moves taken and
remaining goal distance, compared with held-out training-distribution rides. (c) Median true-position
write by goal distance, grouped by moves taken in the full prefix + suffix ride; states above the
give-up threshold are excluded.}
\label{fig:occupancy-compression}
\end{figure}

\begin{figure}[!htb]
\centering
\includegraphics{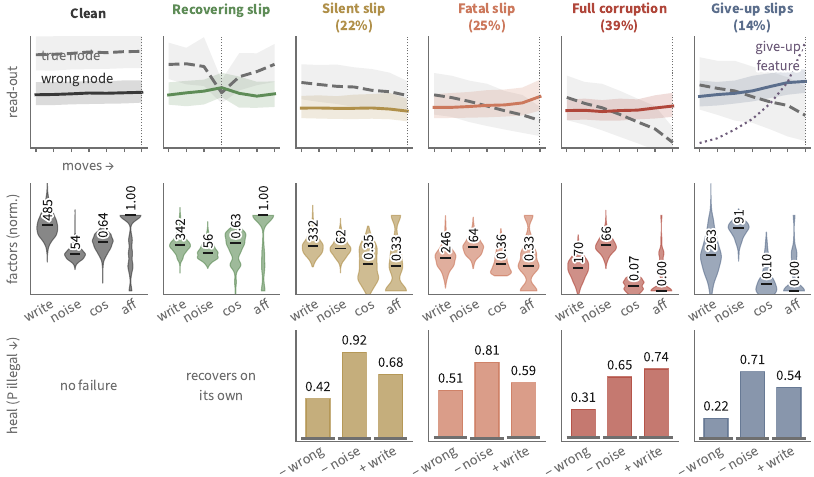}
\caption{\textbf{Compression rides exhibit the same failure modes: weak position writes and noise, with recovery under the same position edits.} Columns compare failure categories and reference states. Rows show position readouts around failure or recovery, scaled write/noise/cosine/affordance measurements, and the fraction of illegal probability removed. Gray marks show matched random edits.}
\label{fig:hero-compression}
\end{figure}

\textbf{The continuations reach challenging states and exhibit the same failure modes.}
The prefix + suffix rides enter the deep-and-far regime, rarely visited in training
(Fig.~\ref{fig:occupancy-compression}b). Within depth bands, the true-position
write weakens with distance to the goal (Fig.~\ref{fig:occupancy-compression}c). About $28\%$
of $21{,}000$ sampled continuations leave the graph. Applying the stress classifier
to their failure states gives $21.7\%$ silent slips, $24.9\%$ fatal slips, $39.4\%$
full corruption, and $13.7\%$ give-up cases (Fig.~\ref{fig:hero-compression}).
Clearing position noise while preserving the true-position write, or restoring
the write, reduces illegal-move probability in every category. Together with the
write and noise measurements, these interventions support the same explanation
as for stress and detour failures.

\textbf{Longer suffixes bring more illegal moves and lower compression.}
To lengthen the suffixes without changing the prefix routes, we move the goal
farther away. We keep two $12$-move routes fixed at each of $60$ intersections
with eligible goals in all seven distance bands, changing only the destination token
(Table~\ref{tab:distance}, lower block). Scores fall from $0.983$ in the nearest
band to $0.167$ in the farthest; illegal moves rise from $0.06$ to $9.18$ per
thousand checked continuation moves. Both prefixes still decode correctly in
$419/420$ pair--goal conditions ($99.8\%$). Across the seven band averages,
compression and illegal-move rate are strongly correlated ($r=-0.957$): the
longer continuations produce more illegal moves and less agreement between
prefixes. This supports the interpretation that compression largely tracks
the same localization failures seen in the stress and detour tests.
For the illegal-move rate, we stop counting at the first illegal move, including
that move. Continuation length includes all generated moves, even those after
an illegal move. The upper block groups the $146$ sampled-prefix trials
by goal distance.

\begin{table}[!htb]
\caption{\textbf{Compression scores fall as the goal becomes more distant.} Upper block: $146$ supplementary trials with sampled prefix lengths. Lower block: the same $60$ pairs of $12$-move routes in each band, changing only the goal (seed~0). Illegal /1k is the off-graph rate per thousand checked moves. Decode requires both prefixes to identify the shared intersection. Dashes indicate unrecorded continuation lengths.}
\label{tab:distance}
\centering\footnotesize
\begin{tabular}{@{}lrrrrrr@{}}
\toprule
& & \multicolumn{2}{c}{median length (moves)} & & & \\
\cmidrule(lr){3-4}
goal distance (moves) & pairs & route & continuation & illegal /1k & diff-means decode & score \\
\midrule
\multicolumn{7}{@{}l}{\emph{sampled-prefix diagnostic run}} \\
\csname @@input\endcsname tables/app_compression_sampled_distance.tex
\multicolumn{7}{@{}l}{\emph{controlled goal-swap experiment: fixed routes, varying destination}} \\
$1$--$4$ & 60 & 12 & 17 & 0.06 & 100\% & 0.983 \\
$5$--$11$ & 60 & 12 & 43 & 0.10 & 100\% & 1.000 \\
$12$--$20$ & 60 & 12 & 62 & 0.11 & 98.3\% & 0.933 \\
$21$--$30$ & 60 & 12 & 72 & 0.27 & 100\% & 0.817 \\
$31$--$45$ & 60 & 12 & 83 & 1.18 & 100\% & 0.700 \\
$46$--$60$ & 60 & 12 & 86 & 4.45 & 100\% & 0.400 \\
$61$--$99$ & 60 & 12 & 85 & 9.18 & 100\% & 0.167 \\
\bottomrule

\end{tabular}
\end{table}

\textbf{Continuous position reinforcement.}
For the reinforcement experiment, we add the true intersection's difference-in-means
vector at layer~11 after each move token ($\alpha=1$), using the large random-walk
model. The layer is selected by the minimal teleportation test. We obtain the true
position by following the actual moves and stop editing after an illegal transition.
For compression, we apply the edit both when generating continuations and when
scoring them under the other prefix. Detour and compression use the original
benchmark implementations. We reuse the $20{,}000$ baseline stress pairs and
evaluate $1{,}000$ detour cases and $1{,}936$ scored compression pairs from
$2{,}500$ attempts (seed~0).

\section{Mechanistic indicators across models and through training}
\label{app:indicators}

\subsection{Across models}
\label{app:models}

\textbf{Feature and layer selection.}
We fit intersection centroids on each model's training rides with a budget of
$100$ states per intersection, evaluate on its held-out rides, and select layers
by that model's sweeps. The decode indicator is the share of intersections read
correctly in at least $90\%$ of their held-out states. Teleportation selects the layer on $200$ scenes, then
reports the share of targets succeeding on more than half of all their eligible
one-move scenes. The original benchmark code supplies compression and
detour scores. Table~\ref{tab:models-full} collects all comparison metrics, including street steering and position
decoding after removing past position codes. For street steering, we inject the source
intersection feature at the preceding position at every layer and sweep all read layers.
Street steering selects the injection-strength
grid and read layer with the best top-one score using fitted response slopes across
strengths. At that choice, the reported top-five accuracy uses the endpoint increase.

\begin{table}[!htb]
\caption{\textbf{World-modeling measurements across models.}
\texttt{SP}, \texttt{NSP} and \texttt{RW} denote shortest paths, noisy shortest paths and random walks;
\texttt{NTP} and \texttt{NextLat} denote next-token and next-latent prediction. Street steering
reports top-five next-intersection accuracy; tracking reports current-position
accuracy before and after removing past position codes. Selected layers appear
below each score.}
\label{tab:models-full}
\centering\footnotesize\setlength{\tabcolsep}{2pt}\renewcommand{\arraystretch}{1.1}
\resizebox{\textwidth}{!}{%
\begin{tabular}{@{}lccccccccc|ccc@{}}
\toprule
& \multicolumn{7}{c}{\textbf{Map / localization}} & \multicolumn{2}{c}{\textbf{Navigation}}
& \multicolumn{3}{|c}{\textbf{Behavior}} \\
\cmidrule(lr){2-8}\cmidrule(lr){9-10}\cmidrule(lr){11-13}
& decode & causal & streets & streets & tracking & super- & by & legal & goal & stress & detour & compr. \\
& & & steer & probe & & position & afford.? & moves & compass & test & test & \\
& {\scriptsize\color{muted}\% int $\ge$.9} & {\scriptsize\color{muted}telep $>$50\%}
& {\scriptsize\color{muted}top5} & {\scriptsize\color{muted}held out}
 & {\scriptsize\color{muted}$-$past pos}
& {\scriptsize\color{muted}angle / dims} & {\scriptsize\color{muted}own class}
& {\scriptsize\color{muted}\% int.} & {\scriptsize\color{muted}steer err}
& {\scriptsize\color{muted}on graph} & {\scriptsize\color{muted}success} & {\scriptsize\color{muted}score} \\
\midrule
\mrow{SP\textperiodcentered NTP}{12L$\times$768d$\times$12h}
 & \cn{\val{12.7\%}{L6}} & \cn{\val{14.9\%}{L10}} & \cn{\val{8.2\%}{L8}} & \cn{\val{23.5\%}{L7}}
 & \cp{\val{.19$\rightarrow$.09}{L6}}
 & \cn{\val{35.9$^\circ$}{162d, L6}} & \cn{\val{41.0\%}{L6}} & \cn{\val{18.1\%}{L10}} & \cp{\val{25.8$^\circ$}{L12}}
 & \cp{71.7\%} & \cn{0.0\%} & \cn{.101} \\
\mrow{NSP\textperiodcentered NTP}{48L$\times$1600d$\times$25h}
 & \cn{\val{17.1\%}{L43}} & \cn{\val{16.9\%}{L46}} & \cn{\val{8.0\%}{L45}} & \cn{\val{12.2\%}{L36}}
 & \cy{\val{.21$\rightarrow$.02}{L43}}
 & \cn{\val{38.0$^\circ$}{251d, L43}} & \cn{\val{41.1\%}{L43}} & \cn{\val{22.3\%}{L44}} & \cp{\val{23.2$^\circ$}{L47}}
 & \cp{74.7\%} & \cn{0.2\%} & \cn{.054} \\
\mrow{RW\textperiodcentered NTP}{48L$\times$1600d$\times$25h}
 & \cy{\val{99.6\%}{L18}} & \cy{\val{94.0\%}{L11}} & \cy{\val{93.5\%}{L17}} & \cy{\val{89.2\%}{L15}}
 & \cy{\val{1.00$\rightarrow$.15}{L18}}
 & \cp{\val{48.1$^\circ$}{244d, L18}} & \cy{\val{91.4\%}{L18}} & \cy{\val{99.4\%}{L35}} & \cy{\val{18.1$^\circ$}{L15}}
 & \cy{91.5\%} & \cp{63.1\%} & \cp{.524} \\
\midrule
\mrow{RW\textperiodcentered NTP}{48L$\times$384d$\times$8h}
 & \cy{\val{99.7\%}{L40}} & \cy{\val{88.5\%}{L31}} & \cp{\val{39.7\%}{L21}} & \cy{\val{81.9\%}{L39}}
 & \cy{\val{1.00$\rightarrow$.11}{L40}}
 & \cp{\val{47.1$^\circ$}{210d, L40}} & \cy{\val{88.2\%}{L40}} & \cy{\val{96.9\%}{L44}} & \cy{\val{17.3$^\circ$}{L43}}
 & \cy{96.7\%} & \cy{77.5\%} & \cp{.523} \\
\mrow{RW\textperiodcentered NextLat}{48L$\times$384d$\times$8h}
 & \cy{\val{99.9\%}{L36}} & \cy{\val{87.1\%}{L27}} & \cn{\val{17.5\%}{L21}} & \cy{\val{94.8\%}{L44}}
 & \cy{\val{1.00$\rightarrow$.17}{L36}}
 & \cp{\val{42.2$^\circ$}{183d, L36}} & \cy{\val{84.7\%}{L36}} & \cy{\val{99.0\%}{L44}} & \cy{\val{17.6$^\circ$}{L37}}
 & \cy{97.3\%} & \cy{78.5\%} & \cp{.556} \\
\bottomrule
\end{tabular}}
\end{table}

\textbf{Street-probe control.}
To check whether low street-probe accuracy reflects difficulty reading the current
position, we also train probes to predict the current intersection. Both tests
use one linear map per move and hold out a quarter of source intersections.
Current-position accuracy
is above $99\%$ on the random-walk models, but only $50.3\%$ and $33.2\%$ on SP and NSP.
For the latter models, even the current position is therefore harder to read
at held-out intersections.

\textbf{Feature-extraction and teleportation controls.}
On the same teleportation scenes, success rises from $3.8\%$ to $26.8\%$ on SP
and from $6.3\%$ to $31.5\%$ on NSP when we apply the position edit. On the
three random-walk models, it rises from below $0.1\%$ to $82.2$--$87.4\%$.
These rates average over scenes, rather than counting the intersections that pass
the test as in Table~\ref{tab:models-full}. We also compare diff-means with probe
directions and whitened means in a separate SP experiment.
The probe gives better held-out decoding than diff-means ($92.3\%$ versus $53.9\%$),
but lower one-move teleportation success at the selected steering layer
($20.5\%$ versus $35.5\%$; $2.5\%$ without an edit). Whitened means give
$69.6\%$ decoding and $4.5\%$ teleportation success. Better decoding alone
therefore does not make a direction better for steering.

\textbf{Legal-move measurement and baseline.}
We normalize probabilities over the eight move tokens and classify moves above
$3\%$ as legal. A prediction is correct only if the complete legal-move set
matches. Always guessing the most common set scores $9.6\%$.
The single-model analysis uses a $1\%$ threshold; the cross-model comparison
and training sweep use $3\%$. The layer sweep excludes the last two layers
of SP and the last four layers of the $48$-layer models; ties use the earliest layer.

\textbf{Position-tracking control.}
We compare removing past position information with removing a random subspace
of the same dimension. Across the five models, random-subspace removal changes
decoding accuracy by at most $0.4$ percentage points, whereas removing past
position information lowers it by $9.6$--$89.0$ percentage points.

\textbf{Training details.}
We use the authors' released checkpoints. Table~\ref{tab:training-exposure}
summarizes their training budgets. \citet{vafa2024worldmodel} train SP until
overfitting and select the best validation checkpoint; for NSP and RW, they
use the last validation checkpoint after five and one epochs, respectively.
\citet{teoh2026nextlat} train for six epochs because performance does not
generally converge within one epoch. The smaller RW models' stronger performance
may therefore partly reflect their longer training.

\begin{table}[!htb]
\centering\small
\caption{\textbf{Reported training budgets of the compared models.} Corpus sizes
and epochs follow \citet{vafa2024worldmodel} and \citet{teoh2026nextlat}.
Total training tokens are corpus tokens multiplied by epochs, counting repeated
passes. SP uses the best validation checkpoint, so its epoch count and total
exposure are unspecified. Token counts are in billions. Global batch sizes count
rides for Vafa models and packed $256$-token sequences for small RW models.}
\label{tab:training-exposure}
\begin{tabular}{@{}lrrrr@{}}
\toprule
Model & Batch size & Corpus tokens (B) & Epochs & Training tokens (B) \\
\midrule
SP & $48$ & $\sim0.12$ & --- & --- \\
NSP & $48$ & $\sim1.68$ & $5$ & $\sim8.39$ \\
Large RW & $48$ & $\sim4.74$ & $1$ & $\sim4.74$ \\
Small RW\textperiodcentered NTP & $256$ & $\sim4.74$ & $6$ & $\sim28.41$ \\
Small RW\textperiodcentered NextLat & $256$ & $\sim4.74$ & $6$ & $\sim28.41$ \\
\bottomrule
\end{tabular}
\end{table}

\textbf{Slip and recovery measurement.}
\label{app:slip-recovery}
For $1{,}500$ stress rides per random-walk model, we rank mean-centered unit
intersection directions at each state, using each model's selected layer. A slip is an incorrect top-ranked node;
recovery means a correct readout within five further moves. If the ride ends before recovery and before five further moves, an illegal
ending counts as a failure to recover; other endings are excluded from the
recovery rate.
Table~\ref{tab:slip-recovery} reports both rates, using the same direction-based
position readout as the failure analyses.

\begin{table}[!htb]
\caption{\textbf{Smaller random-walk models slip less often under stress.}}
\label{tab:slip-recovery}
\centering\small
\begin{tabular}{@{}llrr@{}}
\toprule
model & & slips & recovers \\
& & {\scriptsize\color{muted}\% of steps} & {\scriptsize\color{muted}within 5 steps} \\
\midrule
\texttt{RW\textperiodcentered NTP} & {\scriptsize\color{muted}48L$\times$1600d} & 1.52\% & 67.8\% \\
\texttt{RW\textperiodcentered NTP} & {\scriptsize\color{muted}48L$\times$384d} & 0.56\% & 91.2\% \\
\texttt{RW\textperiodcentered NextLat} & {\scriptsize\color{muted}48L$\times$384d} & 0.24\% & 77.4\% \\
\bottomrule
\end{tabular}

\end{table}

\FloatBarrier
\subsection{Through training}
\label{app:growth}
\textbf{Training and measurement details.}
The model uses \citeauthor{teoh2026nextlat}'s (\citeyear{teoh2026nextlat}) $48$-layer, $384$-wide architecture with eight heads.
We train on random walks with next-token prediction, context length $256$, effective batch size
$256$, seed $1234$, and Adam at learning rate $10^{-4}$ for $300{,}000$ updates.
Each indicator uses the fixed layer shown in
Table~\ref{tab:growth-controls}, selected from the finished run. The table
gives the raw values underlying Fig.~\ref{fig:growth}, with controls, at twelve
checkpoints spanning the run. The figure normalizes each indicator to its final value; for the compass,
we normalize the reduction in angular error from the random baseline of $90^\circ$. We report the share
of intersections decoded at both $90\%$ and $50\%$ accuracy. For street steering,
we inject position features at every layer from $1$ through $24$ and rank next-intersection
features at layer~24. We also measure teleportation without an edit, edit strength
relative to the residual norm, current-intersection probing, and tracking after
random-subspace removal. A different-move street-steering control gives lower
top-five accuracy at every measured checkpoint.

\textbf{The street probing test is a more reliable indicator than the street steering test.}
Between steps $30{,}000$ and $300{,}000$, street steering top-five accuracy falls
from $59.3\%$ to $32.2\%$, while street probing accuracy gradually rises from
$72.4\%$ to $82.6\%$ (Table~\ref{tab:growth-controls}). The latter is more consistent
with improving street encoding. NextLat also performs poorly on the steering test
despite strong probing performance (Table~\ref{tab:models-full}), even though its
training objective encourages learning transitions. These results lead us to prefer
the street probing test as an indicator. We treat strong steering performance as
sufficient, but not necessary, evidence of street encoding. The steering test uses
averaged residual streams and supplies only one previous position, rather than
the look-back window the model normally uses. We hypothesize that changes in how
the model uses past positions during training can make this artificial setup less
effective, even as street connections become more accurately linearly decodable.

\textbf{A high compression score need not imply successful navigation.}
Even a model that navigates poorly can score highly on compression, because both
prefixes can agree on incorrect continuations. At step $200$, compression is
$0.404$, although only $1.8\%$ of stress rides remain on the graph. A high
compression score is therefore informative only when the model also performs reasonably
well on the task. Fig.~\ref{fig:growth} shows compression from the fourth measured
checkpoint (step $1{,}500$); Table~\ref{tab:growth-controls} retains all measurements.

\begingroup
\setlength{\intextsep}{4pt}
\begin{table}[!htb]
\caption{\textbf{World-modeling capacities develop at different stages of training.}
Raw measurements at twelve checkpoints of the small RW\textperiodcentered NTP model, using fixed
layers. The upper block covers localization and street encoding; the lower block covers packing,
navigation, and behavior. Legal-move prediction and the goal compass improve before reliable position decoding.
For causal measurements, floor is teleportation success without an edit and dose
is the edit-to-residual norm ratio. Tracking compares the full history with removal
of past position information ($-$pos) or a random-subspace edit ($-$rand); street
probes predict the next or current intersection.
Packing reports the nearest-neighbor feature angle, the dimensions explaining
$90\%$ of variance ($d_{90}$), and the percentage of features closest to their own
legal-move group's mean (affordance).
Accuracy and success rates are percentages, except causal mean, tracking, and
legal-set accuracy, which are fractions. Compression is also a fraction; angle
and compass error are in degrees. Dashes mark unavailable measurements.}
\label{tab:growth-controls}
\centering\fontsize{8.5}{9.5}\selectfont\setlength{\tabcolsep}{2.5pt}
\begin{tabular}{@{}rrrrrrrrrrrrrr@{}}
\toprule
 & \multicolumn{2}{c}{Decode (L40)} & \multicolumn{4}{c}{Causal (L33)} & \multicolumn{3}{c}{Tracking (L40)} & \multicolumn{2}{c}{\shortstack{Streets: steering\\test (L24)}} & \multicolumn{2}{c}{\shortstack{Streets: probe\\test (L39)}} \\
\cmidrule(lr){2-3}\cmidrule(lr){4-7}\cmidrule(lr){8-10}\cmidrule(lr){11-12}\cmidrule(lr){13-14}
step & $\geq.9$ & $\geq.5$ & $>.5$ & mean & floor & dose & full & $-$pos & $-$rand & top1 & top5 & next & current \\
\midrule
50 & 0.3 & 0.3 & 8.6 & 0.31 & 13.6 & 0.28 & -- & -- & -- & -- & -- & -- & -- \\
200 & 0.4 & 1.4 & 12.4 & 0.37 & 9.2 & 0.64 & 0.03 & 0.02 & 0.04 & 0.6 & 2.0 & 1.6 & 2.2 \\
700 & 3.8 & 20.4 & 26.5 & 0.48 & 10.8 & 1.03 & 0.27 & 0.03 & 0.27 & 7.8 & 18.9 & 12.3 & 22.8 \\
1500 & 8.8 & 43.5 & 32.6 & 0.51 & 10.0 & 1.07 & 0.51 & 0.02 & 0.47 & 10.4 & 25.3 & 20.5 & 43.6 \\
3000 & 21.8 & 80.4 & 50.3 & 0.59 & 6.8 & 1.08 & 0.71 & 0.01 & 0.64 & 10.8 & 27.6 & 34.4 & 65.4 \\
5000 & 40.4 & 96.6 & 57.9 & 0.64 & 3.2 & 1.11 & 0.83 & 0.02 & 0.76 & 13.3 & 32.8 & 46.2 & 78.9 \\
10000 & 71.1 & 99.6 & 53.1 & 0.61 & 0.4 & 1.11 & 0.89 & 0.02 & 0.86 & 24.5 & 52.8 & 59.4 & 90.4 \\
30000 & 95.1 & 100.0 & 66.7 & 0.68 & 0.0 & 1.12 & 0.98 & 0.06 & 0.94 & 28.3 & 59.3 & 72.4 & 97.5 \\
45000 & 97.6 & 100.0 & 66.7 & 0.69 & 0.0 & 1.12 & 0.99 & 0.05 & 0.97 & 25.6 & 56.1 & 74.9 & 98.2 \\
110000 & 99.1 & 100.0 & 77.9 & 0.75 & 0.0 & 1.09 & 1.00 & 0.07 & 0.98 & 19.9 & 46.5 & 79.3 & 98.9 \\
200000 & 99.3 & 100.0 & 82.6 & 0.80 & 0.0 & 1.08 & 1.00 & 0.08 & 0.99 & 10.7 & 28.3 & 81.0 & 99.0 \\
300000 & 99.6 & 100.0 & 88.1 & 0.84 & 0.0 & 1.07 & 1.00 & 0.11 & 0.99 & 12.8 & 32.2 & 82.6 & 99.3 \\
\bottomrule
\end{tabular}
\par\medskip
\begin{tabular}{@{}rrrrrrrrr@{}}
\toprule
 & \multicolumn{3}{c}{Packing (L40)} & \multicolumn{2}{c}{Navigation} & \multicolumn{3}{c}{Behavioral tests} \\
\cmidrule(lr){2-4}\cmidrule(lr){5-6}\cmidrule(lr){7-9}
step & angle & $d_{90}$ & affordance & \shortstack{legal set\\(L44)} & \shortstack{compass\\(L43)} & \shortstack{stress\\on-graph} & \shortstack{detour\\arrival} & compression \\
\midrule
50 & 4.4 & 1 & 32.2 & 0.024 & 86.9 & 0.5 & 0.0 & 0.000 \\
200 & 11.9 & 5 & 45.4 & 0.025 & 77.7 & 1.8 & 0.0 & 0.404 \\
700 & 21.8 & 34 & 62.6 & 0.386 & 32.9 & 2.9 & 0.0 & 0.025 \\
1500 & 25.1 & 76 & 87.1 & 0.868 & 24.6 & 3.3 & 4.0 & 0.000 \\
3000 & 29.3 & 115 & 96.4 & 0.980 & 20.5 & -- & -- & -- \\
5000 & 33.5 & 142 & 97.7 & 0.993 & 20.3 & -- & -- & -- \\
10000 & 38.6 & 167 & 98.3 & 0.994 & 19.4 & 48.5 & 29.0 & 0.081 \\
30000 & 44.3 & 189 & 96.0 & 0.993 & 18.6 & -- & -- & -- \\
45000 & 45.5 & 194 & 95.2 & 0.994 & 19.1 & 87.0 & 59.5 & 0.329 \\
110000 & 46.4 & 199 & 92.2 & 0.992 & 18.3 & -- & -- & -- \\
200000 & 46.5 & 203 & 90.1 & 0.989 & 18.5 & -- & -- & -- \\
300000 & 46.4 & 204 & 89.0 & 0.988 & 18.3 & 95.0 & 77.0 & 0.484 \\
\bottomrule
\end{tabular}

\end{table}
\endgroup

\end{document}